\documentclass[]{fairmeta}
\usepackage{amsthm,amsmath,amssymb}
\usepackage{graphicx}
\usepackage{natbib}
\usepackage{subfig}
\usepackage{booktabs}
\usepackage[table]{xcolor}
\usepackage{bm}
\usepackage{xspace}
\usepackage{tabularx} 

\usepackage{geometry} 
\usepackage{multirow}
\usepackage{enumitem}
\usepackage{fontspec}
\usepackage{xcolor}
\usepackage{sectsty}
\usepackage{subcaption}

\usepackage[hidelinks,breaklinks,colorlinks=true, linkcolor=blue, citecolor=red, urlcolor=cyan]{hyperref}

\newfontfamily\calmregular{Outfit-Regular.ttf}
\newcommand{\calmfont}[1]{{{\calmregular #1}}}

\newtheorem{theorem}{Theorem}[]

\newtheorem{remark1}[theorem]{Remark}

\newcommand{\modelname}{HoloBrain-0\xspace}

\allsectionsfont{\calmfont}

\title{\calmfont{\modelname Technical Report}}

\author{\calmfont{Xuewu Lin}}
\author{\calmfont{Tianwei Lin}}
\author{\calmfont{Yun Du}}
\author{\calmfont{Hongyu Xie}}
\author{\calmfont{Yiwei Jin}}
\author{\calmfont{Jiawei Li}}
\author{\calmfont{Shijie Wu}}
\author{\calmfont{Qingze Wang}}
\author{\calmfont{Mengdi Li}}
\author{\calmfont{Mengao Zhao}}
\author{\calmfont{Ziang Li}}
\author{\calmfont{Chaodong Huang}}
\author{\calmfont{Hongzhe Bi}}
\author{\calmfont{Lichao Huang}}
\author{\calmfont{Zhizhong Su}}

\affiliation{\calmfont{Horizon Robotics}}
\code{\url{github.com/HorizonRobotics/RoboOrchardLab}}
\webpage{\url{horizonrobotics.github.io/robot_lab/holobrain}}
\correspondence{Tianwei Lin at \email{tianwei.lin@horizon.auto}}

\begin{document}

\abstract{

In this work, we introduce \modelname, a comprehensive Vision-Language-Action (VLA) framework that bridges the gap between foundation model research and reliable real-world robot deployment. 
%
%
The core of our system is a novel VLA architecture that explicitly incorporates robot embodiment priors, including multi-view camera parameters and kinematic descriptions (URDF), to enhance 3D spatial reasoning and support diverse embodiments. 
We validate this design through a scalable ``pre-train then post-train" paradigm, achieving state-of-the-art results on simulation benchmarks such as RoboTwin 2.0, LIBERO, and GenieSim, as well as strong results on challenging long-horizon real-world manipulation tasks.
%
Notably, our efficient 0.2B-parameter variant rivals significantly larger baselines, enabling low-latency on-device deployment.
%
To further accelerate research and practical adoption, we fully open-source the entire \modelname ecosystem,
which includes:
(1) powerful pre-trained VLA foundations; (2) post-trained checkpoints for multiple simulation suites and real-world tasks; 
and (3) RoboOrchard, a full-stack VLA infrastructure for data curation, model training and deployment. Together with  standardized data collection protocols, this release provides the community with a complete, reproducible path toward high-performance robotic manipulation.


}

\maketitle


\begin{figure}[htb]
     \centering
     \includegraphics[width=1.0\linewidth]{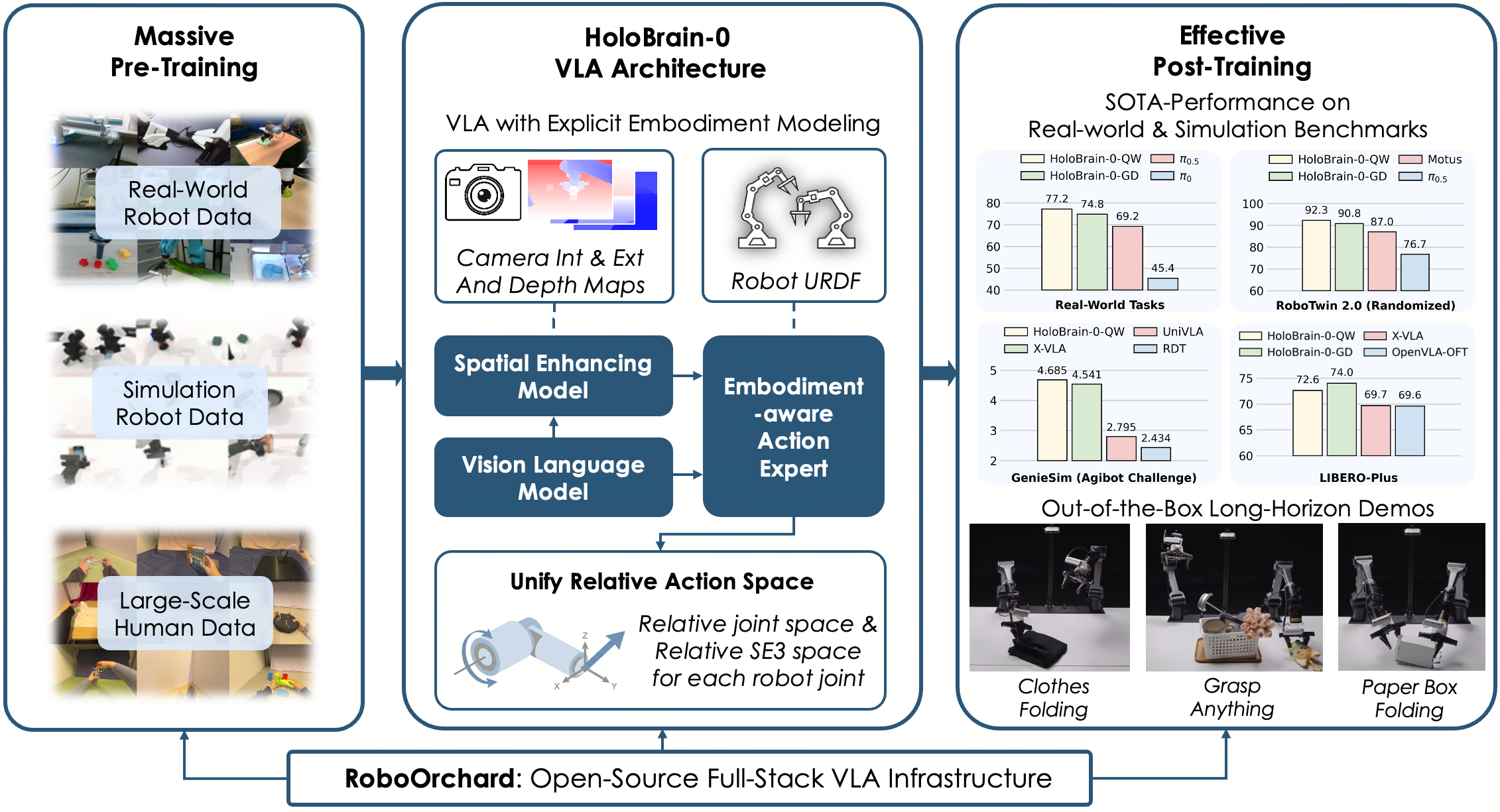}
     \caption{
     Overview of \modelname. 
     %
    %
     By incorporating explicit embodiment modeling (e.g., camera parameters and kinematic descriptions), our model effectively unifies training across heterogeneous robots. 
     Together with a full-stack VLA infrastructure (RoboOrchard) and an effective test-driven data strategy,
     \modelname delivers superior performance on both   real world and simulation manipulation benchmarks.
     }
     \label{fig:overview}
 \end{figure}

\section{Introduction}
\label{intro}

Achieving a truly general-purpose robotic agent has long been a central goal of robotics research. Traditional robotic systems \cite{kaelbling2011hierarchical,garrett2021integrated,guo2023recent} typically rely on modular pipelines that combine perception, state estimation, motion planning, and control. While effective in structured environments, these approaches require extensive task- and platform-specific engineering, which limits their ability to scale and generalize.
Recent progress in robot learning has shifted attention toward end-to-end, data-driven models. Vision–action (VA) models \cite{dp, dp3, peract}, trained via imitation learning, have shown strong capability in learning complex manipulation skills directly from expert demonstrations. Building on this paradigm, the integration of vision–language models (VLM) \cite{beyer2024paligemma,qwen2.5-vl} has led to vision–language–action (VLA) models \cite{kim2024openvla,pi0}.
By combining perception, language understanding, and control, these models pave the way for more general robotic agents.

However, developing a truly general robotic agent remains a significant challenge. 
Although we expect VLAs to understand diverse language instructions, real-world deployment inevitably exposes agents to complex, out-of-distribution (OOD) states that were not seen during training \cite{kim2024openvla,brohan2022rt}. 
This requires robust generalization across various factors, from visual perturbations (e.g., lighting, textures, backgrounds) to physical variations in objects (e.g., poses, shapes, and deformability).
To address these complexities, large-scale pre-training has become  a key strategy \cite{pi0, bjorck2025gr00t,o2024open}. Yet, this introduces a new challenge: utilizing diverse data means handling  different robot embodiments, requiring the model to master cross-embodiment compatibility \cite{o2024open,zheng2025x,wang2024scaling}. 
Building on powerful pre-trained VLA foundations, recent works \cite{zheng2025x,intelligence2025pi,intelligence2025pi0} have demonstrated impressive capabilities in complex, long-horizon tasks via post-training.
%
However, translating these models from isolated demos to robust, general-purpose systems faces systemic bottlenecks. 
Specifically, two critical challenges remain: the prohibitive cost of curating high-quality, expert-level demonstrations at scale, and the difficulty of deploying large models for low-latency, real-time control. 
These challenges underscore a critical reality: scalable progress demands not only advanced architectures but also robust infrastructure. Without a unified stack to streamline data and inference, the path toward generalizability remains bottlenecked by inefficiency and reproducibility issues.

In this work, we present \modelname, a holistic framework that addresses these challenges through a unified design integrating novel VLA model architecture, scalable data strategy, and full-stack infrastructure.

\begin{itemize}[leftmargin=*]

\item \textbf{\calmfont{Embodiment-aware VLA Architecture.} }
Conventional VLAs typically learn a direct mapping from vision to action, relying on composite action spaces \cite{rdt,pi0}, heterogeneous encoder \cite{wang2024scaling} or textual prompts \cite{zheng2025x} to handle data from different robots.
This severely overlooks explicit structural priors, including camera parameters and kinematic chains, forcing models to  fit cross-embodiment data without a physical understanding.
To address this, we explicitly incorporate embodiment priors into our architecture (Fig. \ref{fig:overview}). First, a \textit{Spatial Enhancer} uses camera parameters and depth map to project multi-view images into a unified 3D coordinate system. Second, an \textit{Action Expert} explicitly encodes the robot's kinematic chain via a novel \textit{Joint-Graph Attention} mechanism. Finally, to unify control across heterogeneous robots, we design a hybrid action space that predicts both relative joint and $SE(3)$ motions for each joint.
This ensures compatibility across single/dual-arm manipulators, mobile robots, and human-captured data.

\item \textbf{\calmfont{Effective and Reproducible Data Strategy.}} 
We adopt a two-stage data curation pipeline. For pre-training, we leverage a heterogeneous mixture of multi-embodiment robot and human demonstrations to build a generalist foundation.
For post-training, specifically targeting dexterous and long-horizon bimanual tasks under constrained budgets, we introduce a novel test-driven data collection paradigm. 
This closed-loop approach dynamically adjusts collection strategy based on model performance. 
We detail this iterative process, demonstrating how to selectively target failure cases and ensure high data quality. This provides a reproducible and cost-effective solution for learning complex tasks.

\item \textbf{\calmfont{Open-Source Full-Stack VLA Infrastructure.}} 
Finally, to enable efficient experimentation and deployment, we introduce \textit{RoboOrchard}, an open-source infrastructure that covers the entire pipeline from data collection to model deployment. \textit{RoboOrchard} offers a web-based interface for data acquisition with visualization, ensures data quality through automated validation, organizes data using the MCAP format, and packages datasets into Arrow-based representations for scalable training. For deployment, it supports both synchronous and asynchronous inference, providing flexibility and efficiency across diverse use cases.

\end{itemize}

Driven by this unified design, we extensively evaluate \modelname in both simulation and real-world settings. In simulation, our framework achieves state-of-the-art performance on competitive benchmarks including RoboTwin 2.0, LIBERO, LIBERO-Plus, and GenieSim. In the real world, \modelname  excels in challenging manipulation tasks—such as flexible clothes folding, deformable box folding, and generalized ``Grasp Anything'' tasks. 
Notably, our lightweight 0.2B variant achieves performance comparable to larger baselines, enabling efficient, low-latency on-device deployment.
Finally, we fully open-source the \modelname ecosystem, releasing all pre-trained foundations, post-trained checkpoints, and the \textit{RoboOrchard} infrastructure.

\section{Problem Formulation}

We formulate the robotic manipulation task as a conditional action generation problem. At each time step, the system receives multi-view RGB images $I \in \mathbb{R}^{N\times H\times W \times 3}$ and corresponding depth maps $D \in \mathbb{R}^{N\times H\times W \times 1}$, along with the robot's current proprioceptive joint state $S\in \mathbb{R}^{N_{j}}$. In this work, our policy operates directly on this current observation frame. For multi-task execution, the model is further conditioned on natural language instructions $\mathrm{T}$. Crucially, to explicitly ground the model in the physical world, we incorporate camera parameters $C$ (intrinsics and extrinsics of the $N$ views) and robot kinematic priors $E$ (e.g., URDF descriptions) into the input space. Formally, the complete input space is defined as:

\begin{equation}
    \left\{ I \in \mathbb{R}^{N\times H\times W \times 3},
D \in \mathbb{R}^{N\times H\times W \times 1},
S\in \mathbb{R}^{N_{j}}, \mathrm{T} \right\} \cup \left\{C, E\right\}
\end{equation}

Here, $N$ denotes the number of camera views and $N_j$ is the degrees of freedom (DoF) of the robot. Given these inputs, the model generates action chunk for the subsequent $t_{out}$ steps, denoted as $S_{out}\in \mathbb{R}^{N_{j}\times t_{out}}$.

\section{Method: Model Architecture}
\label{methods_model}


As illustrated in Fig.~\ref{fig:overview}, the overall architecture of \modelname\ presents a unified, end-to-end framework composed of three core modules.
First, a Vision-Language Model (VLM) serves as the semantic backbone, encoding text instructions and visual observations.
To balance inference efficiency with performance, we instantiate this module with two distinct architectures: a 2D detection foundation model (GroundingDINO~\cite{liu2024grounding}) and an LLM-based VLM (Qwen2.5-VL~\cite{qwen2.5-vl}). 
%
Second, a \textit{Spatial Enhancer} (Section~\ref{sec:spatial_enhancer}) injects multi-view spatial consistency and geometric priors into the VLM-extracted visual embeddings.
Third, the \textit{Embodiment-aware Action Expert} (Section~\ref{sec:action_expert}) explicitly encodes the robot's kinematic chain. It facilitates  dense interaction of action queries with robot proprioception, text prompts, and spatially-enhanced visual features via cross-attention mechanisms. Ultimately, it predicts control commands within a unified, relative action space, accommodating heterogeneous robot configurations.
%
Finally, to ensure stable and fluid real-world deployment, we introduce \textit{SimpleRTC} combined with a \textit{Teacher Forcing } training strategy (Section~\ref{sec:simplertc}), enabling smooth asynchronous inference.


\subsection{Perspective-aware Spatial Enhancer }
\label{sec:spatial_enhancer}

To effectively bridge the gap between rich 2D visual semantics and accurate 3D spatial geometry, we adopt the Spatial Enhancer from our prior works (BIP3D \cite{lin2025bip3d} and SEM~\cite{lin2025sem}). This module explicitly injects spatial priors by projecting 2D image features from multiple views along their respective camera frustums into a unified 3D coordinate system. Functionally, it utilizes camera intrinsics and extrinsics to sample 3D points, predicts a discrete depth distribution (with pluggable depth sensor inputs), and aggregates these to generate depth-aware 3D positional embeddings. These embeddings are then fused with image features to produce a globally consistent, geometry-aware 3D representation.
Crucially, to facilitate cross-embodiment training, we modify the original SEM \cite{lin2025sem} design by shifting the 3D projection coordinate frame from the robot's local base frame to the \textbf{central fixed camera frame} (e.g., third-person or head-mounted view).
Fusing auxiliary views (e.g., wrist cameras) into this central frame offers two major advantages: (1) it eliminates learning interference caused by inconsistent base definitions across different robotic platforms, enabling robust cross-embodiment generalization; and (2) it seamlessly accommodates egocentric human data (e.g., EgoDex~\cite{hoque2025egodex}), which inherently lacks a fixed robotic ``base'' frame but possesses a natural head-mounted view.

\begin{figure}[tb]
     \centering
     \includegraphics[width=0.9\linewidth]{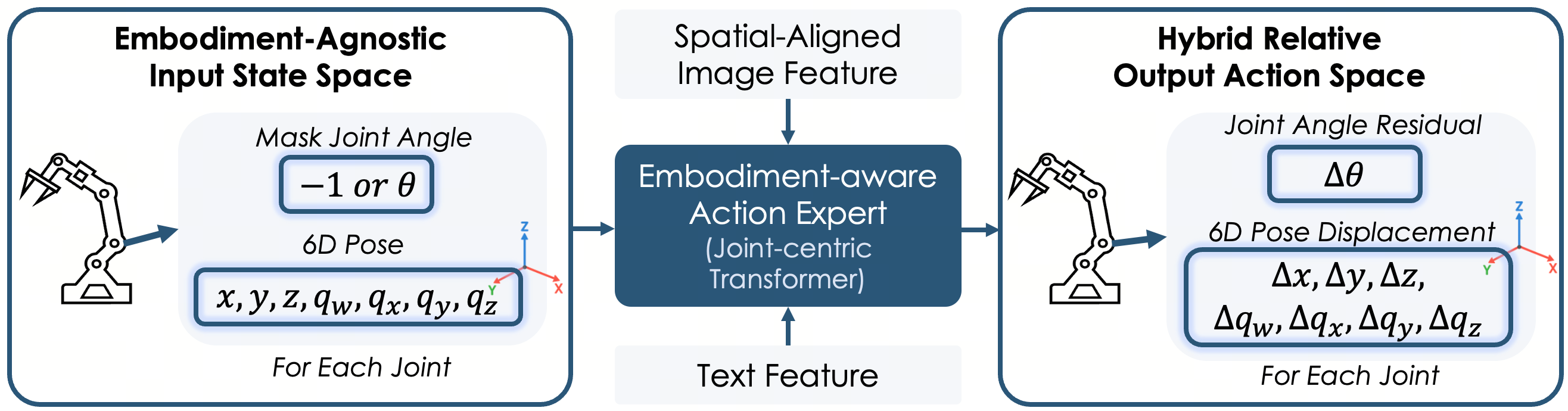}
     \caption{
     Visualization of input state representation and output action space of our action expert. 
     }
     \label{fig:action_space}
 \end{figure}

\subsection{Embodiment-Aware Action Expert}
\label{sec:action_expert}

Motivated by the distinct nature of the robot motion planning task compared to other problem domains, we design an innovative Action Expert, rather than directly adopting the structure of large language models (LLMs) as most existing methods do. The design of the Action Expert adheres to the principle of enhancing embodiment generalization and ensuring compatibility with data from arbitrary embodiment.
The overall architecture largely builds upon  SEM~\cite{lin2025sem}. It consists of a robot state encoder and an action decoder, both primarily implemented with a joint-centric transformer. The core component is the joint-graph attention mechanism (see ~\cite{lin2025sem} for details). To model the action probability distribution, we employ a diffusion-based approach where, instead of predicting noise, we perform x-prediction to directly estimate the states.

Building upon SEM, we introduce improvements to both the input state representation and the output action space, further boosting the model’s embodiment generalization and performance.
First, regarding the input state representation, we mask joint angle information and feed only the 6D pose of each joint into the model.
Formally, the input state for the $i$-th joint is defined as:

\begin{equation}
    s_i = \begin{cases} 
        [-1] \oplus [x, y, z, q_w, q_x, q_y, q_z] & \text{if } m_i = 1 \\
        [\theta] \oplus [x, y, z, q_w, q_x, q_y, q_z] & \text{otherwise}
    \end{cases} \quad 1 \le i \le N_j
\end{equation}

Here, the binary mask $m_i=\text{1}$ indicates that the scalar joint angle $\theta$ is excluded (set to a masked value). 
In our implementation, all joint angles except the gripper openness (in meters) are masked. 
We argue that the joint 6D poses are sufficient to represent a robot’s state. 
Unlike joint angles, which suffer from inconsistent zero-position definitions, rotation directions, and URDF variations across different robot embodiments, Cartesian link poses offer a unified geometric reference. 
%
Consequently, providing ambiguous, embodiment-dependent joint angles as input can hinder the model’s ability to generalize across different embodiments.

For the output action space, our model predicts \textit{hybrid relative transformations} for each joint, encompassing both joint angle space and Cartesian pose space. Specifically, for each joint $i$, the model outputs a concatenation of joint angle residuals (in radians) and link pose displacements (in meters and quaternions):


\begin{equation}
    \mathbf{a}_t = \left\{ 
    \left[ \Delta \theta, \Delta x, \Delta y, \Delta z, \Delta q_w, \Delta q_x, \Delta q_y, \Delta q_z \right]_i 
    \ \middle| \ 1 \le i \le N_j 
    \right\}
\end{equation}

Here, both joint angle and link pose displacements are calculated based on the current robot state, without normalized.
This dual-space prediction formulation offers two significant advantages. First, it enables flexible deployment across diverse hardware interfaces, supporting both low-level joint position control and high-level end-effector pose control. Second, it facilitates training on heterogeneous datasets, including human video data without explicit joint angle annotations.




\subsection{Training Objectives}
\label{sec:loss}

During training, we optimize four loss terms
%
%
as defined in \eqref{eq:loss}.

\begin{equation}
\label{eq:loss}
    L = \alpha_\tau \big( \lambda_1 L_{joint} + \lambda_2 L_{pose} + \lambda_3 L_{pose}^{fk} \big) + \lambda_4 L_{depth}
\end{equation}

Here, the joint position loss $L_{joint}$ and joint pose loss $L_{pose}$ are defined as the distances between the predicted and ground truth joint angles and 6D poses, respectively.
Crucially, as the model predicts relative updates, we add these predictions to the current robot state to compute the loss on absolute values.
To further improve positional accuracy, we add a forward kinematics pose loss $L_{pose}^{fk}$. This loss first recomputes the 6D poses from the predicted joint angles using forward kinematics, and then evaluates the distance between the recomputed poses and ground truth.
Since robot motion trajectories exhibit highly complex distributions and the dataset contains substantial noise, we replace the standard L2 distance used in most diffusion policies with the \textit{smooth L1 distance}, which mitigates instability caused by exceptionally large training errors on certain samples.
The depth loss $L_{depth}$, applied to the depth distribution predicted by the spatial enhancer, is formulated as a cross-entropy loss.
In \eqref{eq:loss}, $\alpha_\tau$ is a timestep-dependent coefficient that varies with the diffusion timestep $\tau$, given by:
\begin{equation}
    \alpha_\tau = T / (\tau + 1)
\end{equation}
where $T$ denotes the maximum number of noise addition steps in diffusion training, set to  $T=1000$ in our experiments.  Consequently, higher noise levels are assigned smaller loss weights, and vice versa.


To further enhance the model's capacity to modeling multimodal action distributions, we introduce a \textit{winner-takes-more} training strategy. 
Specifically, for each training sample, we generate $N$ candidate trajectories and dynamically modulate the loss weights based on prediction error. A higher weight is assigned to the trajectory with the minimum prediction error (the ``winner"), while the weights for the remaining trajectories are suppressed. This mechanism prevents the model from converging to the mean of distinct modes, thereby preserving the diversity of valid solutions.

\subsection{SimpleRTC and Teacher-forcing Training}
\label{sec:simplertc}

In VLA deployment, \textit{synchronous inference} sequentially executes full action chunks. While effective for quasi-static tasks, this modality inherently induces motion pauses and prolonged decision intervals (often reaching $\approx 1$s). Consequently, the system is rendered incapable of handling highly dynamic or fine-grained manipulation tasks. 
Alternatively, \textit{asynchronous inference} decouples model inference from action execution, maximizing inference frequency (typically 5--10 FPS). However, observation latency and the inherent inconsistencies between consecutive action chunks induce trajectory jumps. This results in mechanical jerks that compromise task success rates and potentially damage the hardware.
To mitigate these discontinuities, Black et al. \cite{black2025real} introduced Real-Time Chunking (RTC), which enforces temporal consistency via inference-time gradient guidance from the preceding unexecuted chunk. However, this gradient computation significantly exacerbates inference latency. To address this, they proposed Training-time RTC \cite{black2025training}, incorporating consistency directly into the training phase via fixed-length action prefixes. While eliminating inference overhead, this approach need extensive model retraining.

Conceptually, both approaches draw inspiration from image diffusion inpainting. However, inference-time RTC \cite{black2025real} relies on gradient guidance \cite{song2023pseudoinverse}, which incurs significant computational overhead and optimization instability. Conversely, Training-time RTC \cite{black2025training} functions as a mask-as-condition model, severely restricted by its rigid, fixed-mask structure.
To address these limitations, we rethink the RTC paradigm for VLA models and propose a flexible, two-part strategy:
(1) Inference-time strategy - \textit{SimpleRTC} : A zero-overhead pluggable strategy based on soft-constraint inpainting, seamlessly integrable into any diffusion or flow policy.
(2) Training-time strategy - \textit{Teacher Forcing} . A training strategy that dynamically replaces the first $N$ steps of input noise with ground-truth actions, enabling robust adaptation to arbitrary-length guidance.

\textbf{\calmfont{Inference-time - SimpleRTC.}}
SimpleRTC is a gradient-free method that requires neither model modification nor retraining. SimpleRTC directly leverages the unexecuted segment of the preceding action chunk to guide the denoising phase during inference. 
Adopting a soft-masking strategy inspired by RTC \cite{black2025real}, our method enforces strict consistency with the preceding chunk for the initial $d$ steps (corresponding to the inference latency), followed by a smooth trajectory fusion within a subsequent transition window of length $L$.
Formally, let $\hat{\mathbf{A}}_{0|\tau} = \text{Model}(\mathbf{A}_{\tau}, \tau)$ denote the  x-prediction output at diffusion step $\tau$, and $\mathbf{A}_{\text{prev}}$ represent the unexecuted remainder of the preceding action chunk. The blend action $\tilde{\mathbf{A}}_{0|\tau}$ is formulated as:

\begin{equation}
\tilde{\mathbf{A}}_{0|\tau} = \mathbf{w} \odot \mathbf{A}_{\text{prev}} + (\mathbf{1} - \mathbf{w}) \odot \hat{\mathbf{A}}_{0|\tau}
\label{eq:rtc_main}
\end{equation}

where $\odot$ denotes element-wise multiplication, and $\mathbf{w} \in [0, 1]^H$ is a temporal weighting vector determined by the specific decay strategy.
Then, we conduct denoise sampling as $\mathbf{A}_{\tau-1} = \text{Sampler}(\mathbf{A}_{\tau}, \tilde{\mathbf{A}}_{0|\tau}, \tau) $.
Notably, this guidance is derived for x-prediction. Modifications are necessary for noise or flow prediction parameterizations.
%
To govern this transition, we define a normalized variable $\rho_t$ for each time step $t$ in the prediction horizon $H$, representing the residual influence of the historical trajectory:

\begin{equation}
    \rho_t = 
    \begin{cases}
        1 & t \in [0, d]\\
        1-\frac{t-d}{L}&t \in (d, d+L)\\
        0&t \in [d+L, H)
    \end{cases}
    \label{eq:rtc}
\end{equation}

Here, $d$ represents the inference delay where the trajectory is strictly constrained to history, and $L$ denotes the length of the fusion window.
Experimental results demonstrate that this concise strategy yields surprisingly robust control performance. 
To accommodate varying smoothness requirements across different tasks, we introduce Linear and Quadratic decay curves alongside the original Exponential decay:

\begin{equation}
    w_t^{\text{lin}} = \rho_t, \quad 
    w_t^{\text{quad}} = \rho_t^2, \quad 
    w_t^{\text{exp}} = \rho_t \frac{\mathrm{e}^{\rho_t} - 1}{\mathrm{e} - 1}
    \label{eq:decay_strategies}
\end{equation}

\textbf{\calmfont{Training-time - Teacher Forcing.}}
Since we incorporate inference-time SimpleRTC, the model receives $A$ in \eqref{eq:rtc_main} as input during the inference phase, where the first $t$ steps are already noise-free actions. However, under standard diffusion training, all actions fed to the model are noisy, which leads to a pronounced domain gap between inference and training. To mitigate this discrepancy, we further introduce a \textit{teacher forcing} strategy during training.
Specifically, we construct a hybrid input trajectory by overwriting the initial noisy steps with ground-truth actions.
Let $\mathbf{A}_{gt}$ denotes the ground-truth action, and $\mathbf{A}_{noise}$ denotes the corresponding noisy action. 
We define the teacher-forced input $\mathbf{A}'_{\text{noise}}$ as:
\begin{equation}
\mathbf{A}_{noise,t}' =
\begin{cases}
\mathbf{A}_{gt,t}, & t < N_{\text{prefix}} \\
\mathbf{A}_{noise,t}, & t \ge N_{\text{prefix}}
\end{cases}
\label{eq:a_noise}
\end{equation}

Here, the prefix length $N_{\text{prefix}}$ is sampled from a Poisson distribution $\text{Poisson}(\lambda)$, where the mean $\lambda$ is a hyperparameter calibrated to match the expected inference delay. 
To ensure robustness against history-free inference, we apply this strategy \eqref{eq:a_noise} with a small teacher forcing ratio $\gamma$ (e.g. 25\%). This probabilistic mixture effectively aligns the training distribution with SimpleRTC inference while preserving the model's general denoising capabilities on fully noisy inputs.


%


\section{Method: Data Strategy}

\subsection{Pre-training Data Corpus}

To cultivate a foundational VLA model with robust generalization capabilities, we curate a large-scale, heterogeneous dataset for pre-training. Our data selection strategy is guided by three core principles:

\begin{itemize}[leftmargin=*]

\item \textbf{\calmfont{Cross-Embodiment and Spatial Grounding.}} Aligning with our embodiment-aware architecture, we target extensive cross-embodiment and 3D spatial generalization. Therefore, we source data across diverse robotic platforms and sensor configurations. Crucially, to satisfy our architectural requirements, all selected datasets strictly include multi-view camera parameters (intrinsics and extrinsics) and complete kinematic descriptions (i.e., URDFs).
\item \textbf{\calmfont{High-Fidelity Geometric Priors from Simulation.}} Simulation environments naturally provide exact geometric ground truth, including accurate depth maps and camera parameters. We incorporate a substantial volume of simulation data, as this rich spatial supervision significantly aids the model in developing a rigorous understanding of 3D world dynamics.
\item \textbf{\calmfont{Semantic and Object Diversity.}}
Diversity in task types and interactive objects is critical for open-world generalization. To maximize this diversity, we adopt two explicit data curation strategies. First, we integrate large-scale human-centric dataset, which naturally encapsulate extensive open-world variance in both object appearances and interaction patterns. Second, to systematically benchmark and enhance robotic manipulation across extreme object varieties, we specifically design and collect the ``Grasp Anything" dataset, ensuring the model's robustness against unseen and geometrically complex objects.

\end{itemize}



\begin{figure}
    \centering
    \includegraphics[width=0.9\linewidth]{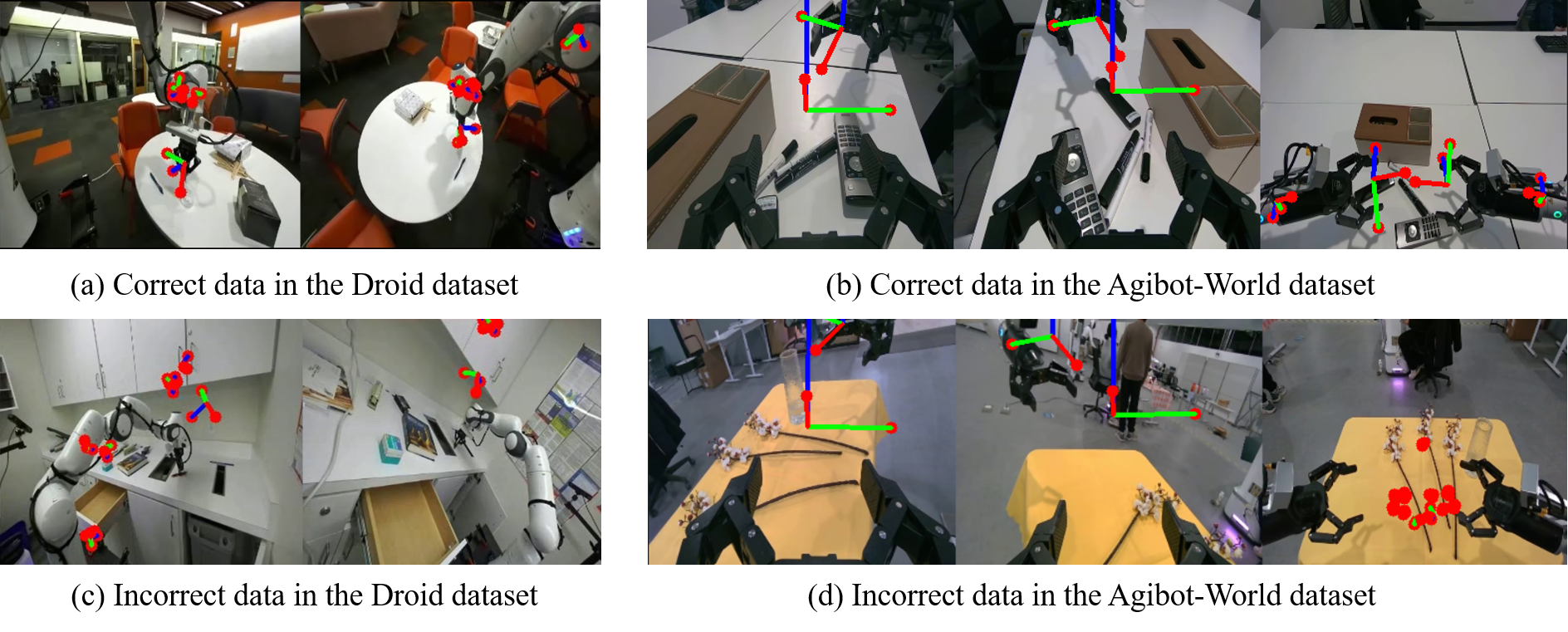}
    \caption{
Visualizing 3D consistency verification. We project the 6D pose of each joint onto the image coordinates (including third-person and wrist cameras) based on the camera intrinsic and extrinsic parameters. Any episodes with inaccurate projection results are identified as erroneous and subsequently filtered out.
    }
    \label{fig:data_quality_inspection}
\end{figure}

After data collection, we conduct rigorous data cleaning. The most critical step is 3D consistency verification: we project joint 6D poses onto images and filter out data exhibiting significant reprojection errors, as shown in Fig.~\ref{fig:data_quality_inspection}. This method enables simultaneous validation of camera parameters, action labels, and URDF accuracy. We also apply further filtering using criteria including task type, motion trajectory plausibility, and instruction-video consistency to eliminate low-quality samples.

As shown in Table~\ref{tab:pretrain_data}, the complete pre-training dataset consists of four main components: proprietary self-collected data, open-source real-world data, simulation data and  human video data. In total, the corpus comprises over 156 million frames (captured over 3,500+ hours) derived from seven distinct embodiments. During pre-training, we manually set sampling ratios for different sources based on data quality and task richness; for instance, we increase the proportion of self-collected data to 41\% and reduce that of the EgoDex dataset from 19\%(frame ratio) to 9.8\%.

\begin{table}[h]
\centering
\caption{Pre-training Data Statistics of \modelname.}
\label{tab:pretrain_data}
\resizebox{\textwidth}{!}{
\begin{tabular}{lllccccc}
\toprule
Data Type & Dataset Name & Embodiment & Frames(M) & Time(H)  & Traj & Training ratio \\
\midrule
\multicolumn{2}{l}{Self-collected}  & Dual-arm Piper & 32.9 & 304.7   & 42539     & 41.41\%   \\
\midrule
\multirow{2}{*}{Real-world} & Agibot World \cite{agibotworld_beta}  & Agibot G1  & 71.6  & 1990.1 & 113846 & 24.5\%  \\
& Droid \cite{khazatsky2024droid}   & Franka  & 14.5 & 267.8 & 54636   & 11.76\%  \\
\midrule
\multirow{5}{*}{Simulation} & \multirow{4}{*}{RoboTwin 2.0 \cite{robotwin2}} & Dual-arm Piper  & 2.65   & 29.5  & 10883  & 4.7\%  \\
& & Dual-arm Franka & 0.34    & 3.7  & 2269    & 0.98\%       \\
& & Dual-arm UR5 & 0.30    & 3.4  & 2032  & 0.98\%       \\
& & Dual-arm Arx & 0.37    & 4.1 & 2306     & 0.98\%       \\
& Agibot-digital \cite{contributors2025agibotdigitalworld} & Agibot G1 & 4.04   & 112.2 & 7350    & 4.9\%        \\
\midrule
Human Video & EgoDex \cite{hoque2025egodex} & Human & 29.9  & 831.9 & 338219 & 9.8\%   \\
\midrule
Total   &    &    & 156.66 & 3547.4    & 531541  & -  \\
\bottomrule
\end{tabular}
}
\end{table}


\subsection{Iterative Test-Driven Data Strategy for Post-Training}

\label{subsec:data_strategy}

Following pre-training, effective post-training is critical for equipping VLAs to master complex, long-horizon tasks. 
However, the efficacy of this phase is often bottlenecked by the high cost of collecting high-quality, real-world data. 
Naive dataset expansion becomes increasingly inefficient for dexterous manipulation due to distribution noise and high marginal collection costs. 
While prior methods address distribution shifts via adversarial collection~\cite{huang2025adversarial, wang2025move} or interactive corrections~\cite{ross2011reduction, wang2025genie}, efficiently scaling \textit{information density} remains a challenge. 
To address this, we propose a test-driven iterative framework that shifts focus from scale-oriented expansion to \textit{quality-driven iteration}. This approach integrates two strategies: proactively expanding state diversity to anticipate potential Out-of-Distribution (OOD) scenarios, and reactively targeting observed failure clusters to align the training distribution with real-world complexities. An empirical analysis of this evolution is provided in Appendix~\ref{supp_sec:sop}.


\textbf{\calmfont{Proactive State Expansion.}} 
To mitigate potential OOD failures before deployment, we proactively enhance state diversity during the data collection phase. 
Rather than relying on random augmentation, we target specific failure-prone conditions across two key dimensions. 
First, regarding visual states, we systematically vary environmental factors such as lighting, background textures, and object instances to ensure perceptual robustness. 
Second, regarding robot states, we transition from full-task demonstrations to granular, sub-task collection. This allows us to specifically reinforce identified bottlenecks in the manipulation pipeline without the redundancy of collecting entire successful trajectories.


\textbf{\calmfont{Test-Driven Failure Recovery.}} 
Inevitably, real-world deployment exposes agents to unanticipated OOD states. 
To close the gap between training-time and test-time distributions, we employ a dynamic, closed-loop recovery strategy.
This process begins with failure mode characterization, where policy failures are systematically analyzed and clustered by root cause. Next, we use these failed states as ``seeds'' for targeted augmentation, varying visual and physical parameters to cover the failure neighborhood. 
Finally, we perform distribution alignment by collecting short-horizon recovery trajectories (typically 2--3s) that guide the agent back to the successful manifold. Unlike standard interactive methods like DAgger~\cite{ross2011reduction, wang2025genie}, which typically address individual errors, our approach resolves entire clusters of failure modes simultaneously, significantly optimizing the data collection budget.

\begin{figure}[h]
    \centering
    \includegraphics[width=1.0\linewidth]{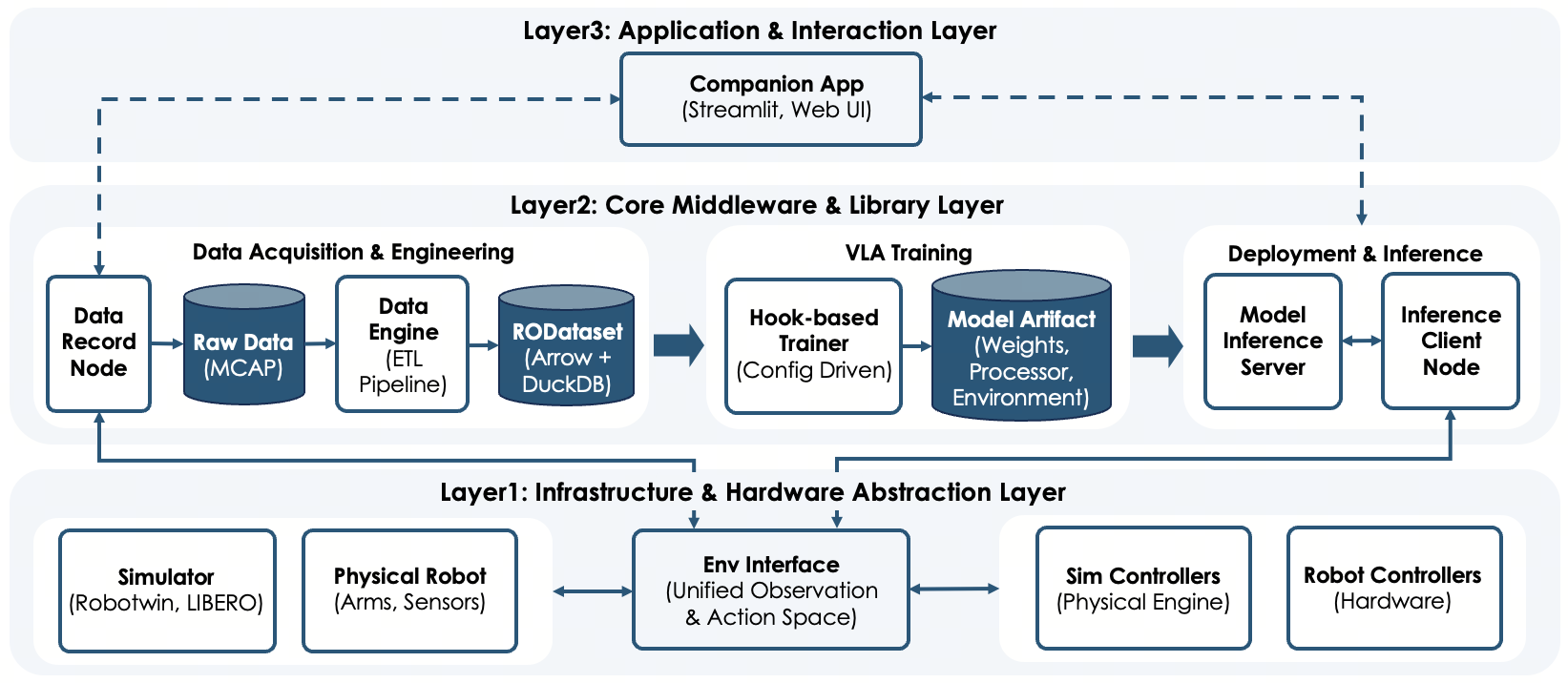} 
    \caption{
    Overview of the RoboOrchard infrastructure. The system comprises three decoupled layers: a bottom \textit{Hardware Abstraction Layer} that bridges the gap between simulation and real-world hardware via unified interfaces; a central \textit{Middleware Layer} that drives the data-to-policy pipeline including storage, training, and deployment; and a top \textit{Interaction Layer} facilitating user management and visualization.
    }
    \label{fig:roboorchard}
\end{figure}

\section{Full-Stack Infrastructure: RoboOrchard}
\label{sec:infrastructure}


%
%

While data-driven imitation learning has received significant attention for robotic manipulation, real-world deployment is often hindered by fragmented toolchains and non-standardized interfaces. 
Unlike the mature field of computer vision, robotic learning lacks unified data standards and deployment paradigms, resulting in substantial engineering overhead. As a consequence, researchers often expend substantial effort on system integration tasks rather than focusing on algorithmic innovation.

To address these challenges, we introduce \textbf{RoboOrchard}, a full-stack modular infrastructure illustrated in Fig.~\ref{fig:roboorchard}. 
Its design centers on the principle of \textit{Artifact-Driven Decoupling}: rather than relying on tightly coupled code across system layers, RoboOrchard links independent modules through standardized artifacts. 
To support this, we define \textit{Unified Artifact Specifications} that precisely describe the protocols for key assets, including self-describing datasets and self-contained model artifacts.
%
Building on these specifications, RoboOrchard provides a suite of ready-to-use components, such as a companion app for high-fidelity data collection (Sec~\ref{sec:orchard_data}), a Pydantic-based training framework (Sec~\ref{sec:orchard_train}), and a deployment runtime that smoothly switches between simulation and real hardware (Sec~\ref{sec:orchard_deploy}). 
Additionally, we provide several convenient tools, such as an interactive visual hand-eye calibration node and a lightweight file server for data review.
%
%
Together, these components reduce engineering effort and accelerate experimental iteration on robot systems.

\subsection{Infrastructure for Data Acquisition}
\label{sec:orchard_data}

High-quality real-world data serves as the cornerstone of embodied intelligence. RoboOrchard establishes a complete infrastructure ranging from hardware acquisition to standardized storage, aimed at resolving issues regarding the temporal synchronization and storage efficiency of multimodal data.

\textbf{\calmfont{Data Acquisition System.}}
We provide a unified data collection suite comprising a high-performance \textit{ROS 2 Recorder} backend and a user-friendly \textit{Companion App}. 
To guarantee high-bandwidth write performance, the backend leverages the \textbf{MCAP} format, featuring an integrated \textit{Integrity Monitor} that automatically flags sensor frame drops or timestamp anomalies. Complementing this, the frontend application offers a ``What You See Is What You Get'' visual interface for task orchestration (via Foxglove). This interactive workflow significantly lowers the barrier to entry for collecting expert demonstrations in laboratory settings.


\textbf{\calmfont{RODataset Specification.}}
%
To overcome the limitations of traditional flat files in handling large-scale multimodal data, we propose the \textit{RODataset Specification}. 
This specification adopts a hybrid storage architecture: high-bandwidth sensor data (e.g., images, joint states) is  stored in the \textit{Apache Arrow} format, 
enabling zero-copy memory loading and seamless integration with the HuggingFace \cite{wolf2020transformers} ecosystem.
Meanwhile, global metadata such as task descriptions and statistical metrics are 
structured via an embedded \textit{DuckDB} engine.
This design enables efficient SQL-based querying on million-frame datasets and natively handles non-aligned multi-frequency signals, preserving the authentic temporal dynamics of the physical world.

\subsection{Infrastructure for Model Training}
\label{sec:orchard_train}


As a core submodule of RoboOrchard, \textit{RoboOrchardLab} is a modular training framework built on a ``Library-First'' philosophy, designed to enhance the reproducibility of algorithm research and the safety of configuration management.
\textit{RoboOrchardLab} currently provides native support for multiple VLA models, such as \modelname, with plans to integrate additional models and learning algorithms in future releases.

\textbf{\calmfont{Type-Safe Configuration \& Hook-based Trainer.}}
To move beyond traditional error-prone, unstructured dictionary configurations, we employ a strict schema-based configuration system powered by \textit{Pydantic}. 
This approach ensures runtime automatic structural validation and enables IDE autocompletion for complex hyperparameters, effectively mitigating potential failures caused by configuration mismatches at the source. 
Furthermore, the training pipeline adopts a hook-based, event-driven architecture. This design empowers researchers to inject custom monitoring, logging, or control logic at specific execution stages without modifying the core training loop, ensuring both flexibility and codebase integrity.


%

\textbf{\calmfont{Model Zoo \& Model Artifacts.}}
The framework provides a baseline algorithm library (\textit{Model Zoo}) that adheres to unified I/O interfaces. 
%
Crucially, once training is completed, the system automatically packages a standardized \textit{Model Artifact}. Unlike simple weight files, this artifact encapsulates \textit{SafeTensors} model weights together with preprocessing pipelines, and environment dependencies into a single portable unit. 
This \textit{self-contained} design ensures that the model maintains independent inference capabilities after being detached from the training codebase, embodying the principle of ``Training as Delivery''.

\subsection{Infrastructure for Model Deployment}
\label{sec:orchard_deploy}

In the deployment phase, RoboOrchard prioritizes enhancing both the reliability and efficiency of the workflow. 
We achieve this through three key pillars: decoupled inference and control nodes with standardized interfaces, flexible inference modes, and intuitive visualization tools. 
Collectively, these designs also enable a seamless transition between real-world robots and simulation environments.


\textbf{\calmfont{Decoupled Inference Architecture.}}
Prevalent open-source deployment scripts often entangle model initialization with robot control loops. This coupling leads to severe dependency conflicts (e.g., between model libraries and hardware drivers) and hinders model switching. 
%
To resolve this, RoboOrchard implements a decoupled \textit{Client-Server Architecture}.
On the server side, we provide a \textit{Flask-based Inference Service}. It can directly load standardized model bundle without modification and can be deployed locally or remotely to isolate environments. This single service unifies inference for both simulation and physical hardware.
On the robot client side, we provide a lightweight \textit{Generic ROS2 Client}. By reading the metadata from the model bundle, this node dynamically configures its subscription (observation) and publication (action) topics, formatting raw sensor data into standardized requests for the inference server.

\textbf{\calmfont{Inference Modes.}}
To accommodate diverse task requirements, the ROS2 client supports two distinct inference modes:
(1) Synchronous mode: Enforces strict ``Observation-Inference-Execution'' timing. It is suited for model debugging and quasi-static manipulation tasks.
(2) Asynchronous mode: Decouples perception and inference frequencies via non-blocking callbacks and timestamp alignment. This significantly improves responsiveness in highly dynamic tasks. 
%
Crucially, the client feeds back execution latency and unexecuted action history to the model server. This context enables server-side smoothing algorithms, such as \textit{SimpleRTC}, to generate temporally consistent action chunks that compensate for asynchronous delays.


\textbf{\calmfont{Interactive Evaluation Interface.}}
To close the loop between model evaluation and data expansion, we seamlessly integrated the deployment runtime into the data acquisition \textit{Companion App}. 
%
%
This unified interface provides centralized control over both the model inference server and the ROS2 Robot Client.
Specifically, users can easily manage the server lifecycle (e.g., specifying model paths and launching services) while simultaneously commanding the client execution flow via fine-grained controls (e.g., \texttt{Start}, \texttt{Pause}, and \texttt{Takeover}). 
%
%
Coupled with real-time recording, this design transforms the deployment tool into a powerful data engine, enabling the efficient collection of both automatic model rollouts and human intervention data, thereby accelerating the iterative improvement of policies.
See Appendix xx for more details of companion App.

\section{Experiments }

In this section, we conduct extensive experiments to evaluate the effectiveness of our proposed \modelname model. Our evaluation is two-fold: First, we benchmark our method against state-of-the-art (SOTA) baselines across 10 diverse real-world manipulation tasks and 4 representative simulation benchmarks. 
Second, we present a detailed analysis of key experiments that offer valuable insights, including how multi-task co-training improves success rates, and how our proposed joint training and inference-time strategies enhance the smoothness and performance of asynchronous inference in dexterous tasks.
%


\subsection{Implementation Details}

%
We adopt Grounding DINO Tiny~\cite{liu2024grounding} and Qwen2.5-VL-3B~\cite{qwen2.5-vl} as our VLM backbones. For Qwen2.5-VL-3B, we freeze both the vision and text encoders to preserve their pre-trained capability to extract semantic representations. Importantly, for the LLM component, we retain only the first transformer layer and discard all subsequent layers to reduce computational overhead. 
%
The resulting parameter statistics are detailed in Tab.~\ref{tab:param}, with \modelname -QW containing 1.1B parameters and \modelname -GD containing  0.2B.
Compared with prevalent VLA models such as GR00T~\cite{bjorck2025gr00t}, $\pi_0$~\cite{pi0}, and RDT~\cite{rdt}, \modelname features a significantly more lightweight action expert. 
This aligns with a hierarchical design philosophy: the VLM backbone acts as the ``brain'' for semantic reasoning, while the action decoder is solely responsible for execution. This focused role allows the action expert to require significantly fewer parameters, resulting in a lightweight architecture that enables efficient deployment on resource-constrained edge devices (e.g., RDK S100~\cite{drobotics_rdks100}).

 \begin{table}[th]
        \caption{
    Model parameter counts (in millions). GD denotes GroundingDINO Tiny, QW represents Qwen2.5-VL-3B.
    LM refers to the language model component. In GD, the LM consists of a BERT module and a feature enhancer. For QW, we retain only the first layer of its LM, resulting in a reduced parameter size of 388.24M. We freeze the BERT module in GD and the vision encoder in QW during training.
    }
    \centering
    \begin{tabular}{l|cccc|cc}
    \toprule
    Model & Vision Encoder & LM & Spatial Enhancer  &  Action Expert & Trainable & All \\
    \midrule
    \modelname -GD &   29.64 & 130.80 & 2.28 & 20.79 &  74.81 &  \textbf{183.70}\\
    \modelname -QW & 668.68 & 388.24 &  2.09  & 20.79 & 412.17 & \textbf{1080.86}\\
    \bottomrule
    \end{tabular}

    \label{tab:param}
\end{table}

\textbf{\calmfont{Training Details.}}
During the pretraining stage, both \modelname-QW and \modelname-GD are trained for 200k steps, with batch sizes of 2048 and 512, respectively.
%
The learning rate remains fixed at $1×10^{−4}$
throughout the entire pretraining phase.
In the post-training stage, the number of training steps is selected within the range of 100k–200k depending on the dataset size. The learning rate starts at $1×10^{−4}$
and is decayed to $1×10^{−5}$ during the final 10\% of the training steps. The batch sizes for post-training are set to 256 for HoloBrain-0-QW and 128 for HoloBrain-0-GD.
The AdamW optimizer is used consistently across all stages.
Refer to Appendix \ref{supp_sec:model} for more implementation details.

%


%
%

\subsection{Results on Real-world Tasks}

\begin{figure}[tb]
     \centering
     \includegraphics[width=1\linewidth]{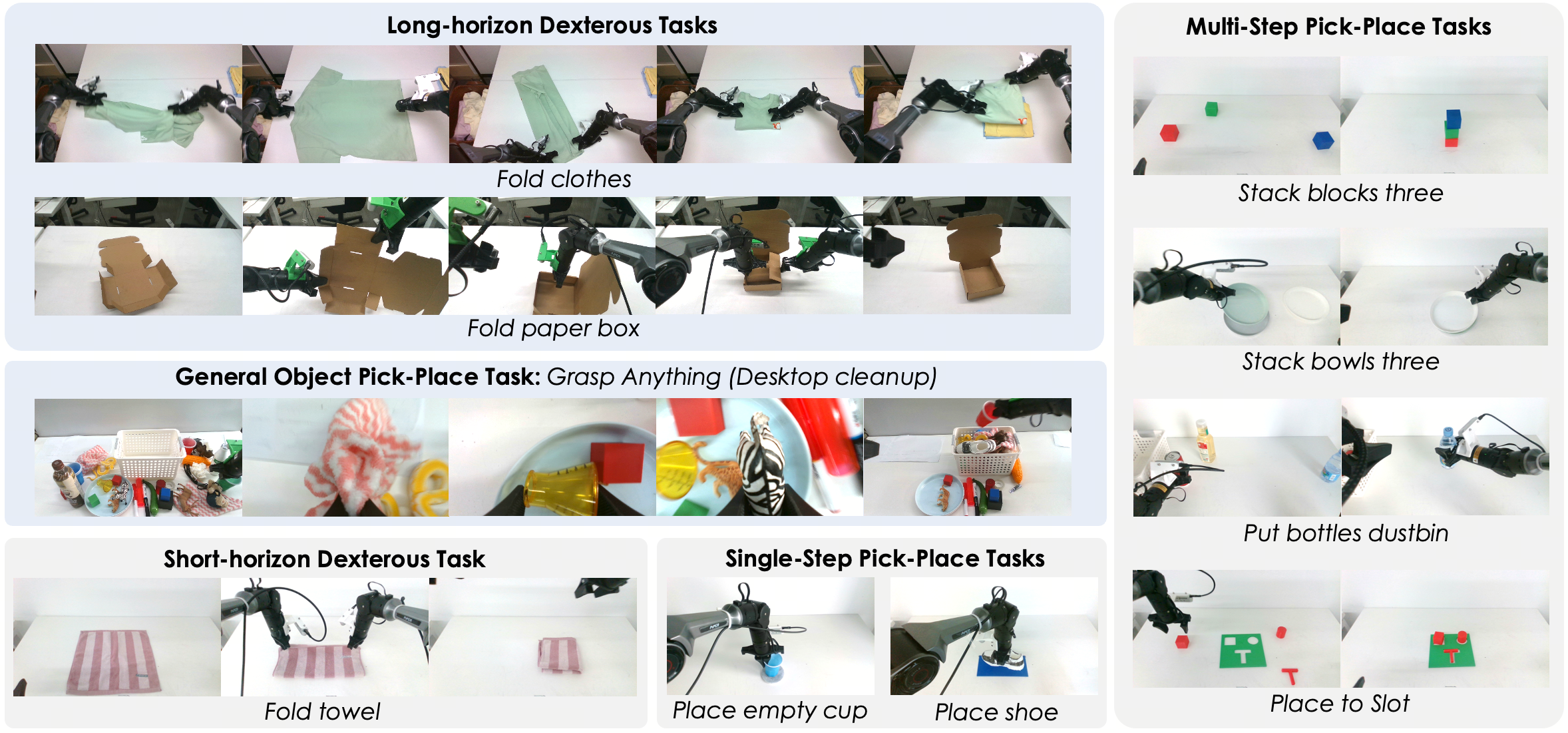}
     
     \caption{
     Real-world evaluation task suite for \modelname. The suite comprises 7 basic tasks (shaded in gray), 2 long-horizon dexterous manipulation tasks, and 1 general object pick-and-place task.
     }
     \label{fig:realword_pic}
 \end{figure}

\textbf{\calmfont{Real World Task Setup.}}
We evaluate our models on the \textit{Dual-arm Piper} embodiment setting (Fig. \ref{fig:robot_setup}).
As illustrated in Fig.~\ref{fig:realword_pic}, we designed a comprehensive suite of real-world experiments categorized into two primary groups to rigorously evaluate our model.
The first category consists of 7 fundamental tasks (short-horizon dexterous, single-step, and multi-step pick-and-place). 
We collect only 200 episodes per task and train a single multi-task model, aiming to validate the model's learning efficacy with limited post-training data.
The second category targets challenging, long-horizon tasks, including \textit{cloth folding}, \textit{box folding}, and \textit{grasp anything}. The first two tasks involve the long-horizon manipulation of deformable objects, representing a research direction currently attracting significant attention. 
In contrast, the grasp anything task emphasizes object diversity. Conditioned on a single high-level instruction (e.g., ``cleanup desktop"), the model must clear a wide variety of objects from a table into a basket. During evaluation, the object set is evenly divided between seen and unseen items.
Utilizing a dataset of approximately 30 hours per task, we adopt a test-driven paradigm to validate the synergy between our model and the data collection pipeline.

%
%


For quantitative evaluation, we measure the success rate and task-specific progress score (detailed in Appendix~\ref{supp_sec:real}) over average 20 trials per task. Crucially, to guarantee a fair comparison while covering a diverse range of initial states, we strictly utilize a fixed set of pre-defined object poses for every evaluation rollout. Specifically, to ensure evaluation consistency, the fold clothes task is evaluated starting from Step 3 (Adjusting); the complete pipeline is illustrated in Fig.~\ref{fig:task_sop}.

\begin{table}[h]
    \centering
        \caption{Real-world robot experiment results. Each cell follows the format ``progress score / success rate''.}
    \resizebox{\textwidth}{!}{
    \begin{tabular}{l|cccccc}
    \toprule
        model & Fold towel & Place empty cup & Place shoe & Stack blocks three & Stack bowls three & Put bottles basket\\
    \midrule
        $\pi_{0}$ & 61.58 / 31.58 & 48.5 / 30.00 & 75.48 / 48.39 & 70.00 / 13.33 & 97.78 / 93.33 & 55.24 / 21.43 \\
        $\pi_{0.5}$ & 61.58 / 63.16 & \textbf{99.50} / \textbf{95.00} & 84.19 / 54.84 & 80.00 / 26.67 & \textbf{100.00} / \textbf{100.00} & 94.29 / 78.57 \\
        HB-GD & \textbf{95.26} / \textbf{84.21} & 89.50 / 85.00 & \textbf{96.77} / \textbf{96.77} & 81.11 / 40.00 & 98.22 / 80.00 & \textbf{96.67} / 78.57 \\
        HB-QW & 84.74 / \textbf{84.21} & 78.50 / 70.00 & 93.23 / 93.55 & \textbf{83.33} / \textbf{46.67} & \textbf{100.00} / \textbf{100.00} & 95.24 / \textbf{85.71} \\

    \toprule
        model & Place to slot & Grasp anything & Fold clothes & Fold paper box & Average progress & Average success rate\\
    \midrule
        $\pi_{0}$ &  68.00 / 33.33 & 87.50 / 87.50  & 33.33 / 15.00 &  86.00 / 80.00 & 68.34 & 45.39 \\
        $\pi_{0.5}$ &  \textbf{84.22} /\textbf{60.00} & \textbf{98.40} / \textbf{98.40} & 60.95 / 50.00 & 81.50 / 65.00 & 84.46 & 69.16\\
        HB-GD  & 80.00 / \textbf{60.00} & 93.50 / 93.50 & 67.62 / 55.00 & 82.00 / 75.00 & \textbf{88.07} & 74.81 \\
        HB-QW & 62.22 / 26.67 & 95.00 / 95.00 & \textbf{81.43} / \textbf{75.00} & \textbf{99.50} / \textbf{95.00} &  87.32 & \textbf{77.18} \\
    \bottomrule
    \end{tabular}
    }

    \label{tab:real_world}
\end{table}

\textbf{\calmfont{Results Analysis.}}
We compare \modelname with $\pi_0$ and $\pi_{0.5}$, and the results are summarized in Table~\ref{tab:real_world}. As a baseline, $\pi_0$ achieves a progress score of 68.34 and a success rate of 45.39\%. Compared to $\pi_0$, $\pi_{0.5}$ makes only minor modifications to the model architecture but undergoes more extensive and better-designed pretraining; its real-world performance improves significantly, particularly excelling in pick-and-place tasks such as \textit{place empty cup} and \textit{grasp anything}.
Compare with $\pi_{0.5}$, \modelname achieves superior performance: \modelname-GD and \modelname-QW outperform $\pi_{0.5}$ by average success rates of \textbf{5.65\%} and \textbf{8.02\%}, respectively, across the 10 evaluated tasks. Further experimental findings are summarized as follows:

\begin{itemize} [leftmargin=*]

\item  \modelname substantially outperforms $\pi_{0.5}$ on two long-horizon tasks, \textit{fold clothes} and \textit{fold paper box}, with success rates improved by 25\% and 30\%, respectively.

\item  In scenarios with fewer stages and limited data, \modelname-GD shows certain advantages over the larger \modelname-QW; however, for long-horizon complex tasks, \modelname-QW is recommended.

\item Benefiting from our data acquisition strategy — the iterative test-driven data strategy (Sec.~\ref{subsec:data_strategy}), we achieve highly robust performance on long-horizon dexterous tasks using only around 30 hours of data, and the system can operate continuously for extended periods without getting stuck.


\item For the \textit{grasp anything} task, Table~\ref{tab:real_world} reports the average success rates across both seen and unseen object categories. Specifically, \modelname -QW achieves 93.5\% on seen objects and 97.5\% on unseen objects. We attribute the slightly lower performance on seen objects to the inherent difficulty of the test set, which includes several objects with challenging, irregular geometries. Crucially, these results indicate that grasping success is largely independent of whether an object appeared during training, demonstrating that our data collection strategy effectively equips the model with generalized grasping capabilities.

\end{itemize}


\subsection{Results on Simulation Benchmarks}

Although the sim-to-real gap persists, simulation benchmarks remain essential for evaluating a model's capacity to fit specific task distributions under fixed data budgets. 
Furthermore, certain benchmarks are designed to target and evaluate specific axes of generalization.
To provide a holistic evaluation, we utilize four benchmarks—RoboTwin 2.0~\cite{robotwin2}, LIBERO~\cite{liu2023libero}, LIBERO-plus~\cite{fei2025libero}, and GenieSim Digit World~\cite{contributors2025agibotdigitalworld}—thereby assessing our method's performance across a spectrum of tasks and environments.

\begin{table}[t]
\centering
\caption{
Evaluation on the RoboTwin2.0 benchmarks (Clean vs Randomized, 50 tasks). 
The performance metrics for $\pi_{0.5}$~\cite{intelligence2025pi0} are cited from Lingbot-VLA~\cite{wu2026pragmatic}, while the results for X-VLA~\cite{zheng2025x} are sourced from Motus~\cite{bi2025motus}. See Table~\ref{tab:robotwin_detail_gd} and Table~\ref{tab:robotwin_detail_qw} for our full results.
}
\label{tab:robotwin_main}
\resizebox{\textwidth}{!}{%
\begin{tabular}{lcccccccccccc}
\toprule
\multicolumn{1}{c}{\multirow{3}{*}{\textbf{Simulation Task}}} & \multicolumn{2}{c}{$\pi_{0.5}$} & \multicolumn{2}{c}{{X-VLA}} & \multicolumn{2}{c}{Lingbot-VLA} & \multicolumn{2}{c}{{Motus}} & \multicolumn{2}{c}{{\modelname-GD}} & \multicolumn{2}{c}{{\modelname-QW}} \\
 & \multicolumn{2}{c}{\textit{(3B)}} & \multicolumn{2}{c}{\textit{(0.9B)}} & \multicolumn{2}{c}{\textit{(4B)}} & \multicolumn{2}{c}{\textit{(8B)}} & \multicolumn{2}{c}{\textit{(0.2B)}} & \multicolumn{2}{c}{\textit{(1.1B)}} \\
\cmidrule(lr){2-3} \cmidrule(lr){4-5} \cmidrule(lr){6-7} \cmidrule(lr){8-9} \cmidrule(lr){10-11} \cmidrule(lr){12-13}
 & Clean & Rand. & Clean & Rand. & Clean & Rand. & Clean & Rand. & Clean & Rand. & Clean & Rand.\\
\midrule
\textit{Place Dual Shoes}    & 75\% & 75\% & 79\% & 88\% & 87\% & 86\% & 93\% & 87\% & 92\% & 95\% & \textbf{96\%} & \textbf{96\%} \\
\textit{Move Stapler Pad}    & 56\% & 42\% & 78\% & 73\% & 74\% & 48\% & \textbf{83\%} & \textbf{85\%} & 74\% & 77\% & 73\% & 84\% \\
\textit{Stack Blocks Two}    & 97\% & \textbf{100\%} & 92\% & 87\% & \textbf{100\%} & 99\% & \textbf{100\%} & 98\% & \textbf{100\%} & 97\% & \textbf{100\%} & 99\% \\
\textit{Scan Object}         & 72\% & 65\% & 14\% & 36\% & \textbf{92\%} & \textbf{96\%} & 67\% & 66\% & 85\% & 80\% & 84\% & 84\% \\
\textit{Place Object Stand}  & 91\% & 85\% & 86\% & 88\% & 93\% & 88\% & \textbf{98\%} & \textbf{97\%} & 92\% & 93\% & 94\% & 94\% \\
\textit{Place Fan}           & 87\% & 85\% & 80\% & 75\% & 92\% & 87\% & 91\% & 87\% & 90\% & 92\% & \textbf{96\%} & \textbf{92\%} \\
\textit{Move Pillbottle Pad} & 84\% & 61\% & 73\% & 71\% & 92\% & 90\% & 93\% & 96\% & \textbf{96\%} & \textbf{97\%} & 95\% & 91\% \\
\textit{Pick Dual Bottles}   & 93\% & 63\% & 47\% & 36\% & \textbf{99\%} & 90\% & 96\% & 90\% & 98\% & \textbf{97\%} & 98\% & 96\% \\
\textit{Blocks Ranking Rgb}  & 92\% & 85\% & 83\% & 83\% & 92\% & 91\% & 99\% & 97\% & \textbf{100\%} & \textbf{98\%} & 98\% & 96\% \\
\multicolumn{1}{c}{......(50 tasks)} & & & & & & & & & & & & \\
\textit{Turn Switch}         & 62\% & 54\% & 40\% & 61\% & 67\% & 63\% & 84\% & 78\% & 70\% & 67\% & \textbf{89\%} & \textbf{79\%} \\
\textit{Pick Diverse Bottles}& 81\% & 71\% & 58\% & 36\% & 88\% & 85\% & \textbf{90\%} & \textbf{91\%} & 86\% & 87\% & 89\% & 87\% \\
\textit{Place Bread Basket}  & 77\% & 64\% & 81\% & 71\% & 95\% & 93\% & 91\% & 94\% & \textbf{97\%} & \textbf{95\%} & 93\% & 90\% \\
\textit{Stack Blocks Three}  & 91\% & 76\% & 6\% & 10\% & 96\% & 95\% & 91\% & 95\% & \textbf{96\%} & \textbf{98\%} & 93\% & 94\% \\
\textit{Put Bottles Dustbin} & 84\% & 79\% & 74\% & 77\% & 92\% & 93\% & 81\% & 79\% & 93\% & 97\% & \textbf{97\%} & \textbf{98\%} \\
\textit{Place Can Basket}    & 62\% & 62\% & 49\% & 52\% & 75\% & 72\% & 81\% & 76\% & \textbf{89\%} & 80\% & 79\% & \textbf{90\%} \\
\textit{Stamp Seal}          & 79\% & 55\% & 76\% & 82\% & 74\% & 77\% & \textbf{93\%} & \textbf{92\%} & 75\% & 76\% & 77\% & 85\% \\
\textit{Hanging Mug}         & 18\% & 17\% & 23\% & 27\% & 34\% & \textbf{53\%} & 38\% & 38\% & 48\% & 45\% & \textbf{55\%} & 52\% \\
\textit{Handover Block}      & 66\% & 57\% & 73\% & 37\% & 83\% & 95\% & 86\% & 73\% & 96\% & 93\% & \textbf{100\%} & 90\% \\
\textit{Stack Bowls Three}   & 77\% & 71\% & 76\% & 86\% & 71\% & 77\% & 79\% & 87\% & \textbf{92\%} & 81\% & 88\% & \textbf{88\%} \\
\textit{Place Object Basket} & 80\% & 76\% & 44\% & 39\% & 90\% & 88\% & 81\% & 87\% & 87\% & 81\% & \textbf{92\%} & \textbf{90\%} \\
\textit{Open Microwave}      & 34\% & 77\% & 79\% & 71\% & 91\% & 92\% & 95\% & 91\% & \textbf{99\%} & 98\% & 97\% & \textbf{99\%} \\
\midrule
\textbf{Average (\%)}        & 82.74 & 76.76 & 72.80 & 72.84 & 88.56 & 86.68 & 88.66 & 87.02 & 91.30 & 90.80 & \textbf{91.90} & \textbf{92.30} \\
\bottomrule
\end{tabular}%
}
\end{table}

\textbf{\calmfont{RoboTwin 2.0 Benchmark.}}
%
We train a 50-task multi-task model. We utilize a dataset mix of 50 clean and 500 randomized demonstrations per task, where randomization includes background, lighting, table height, and cluttered table. The model undergoes 200k steps of post-training. For comprehensive evaluation, we execute 100 trials per task in both clean and randomized settings aggregating to 10,000 total rollouts.
On this benchmark, our approach achieves SOTA-level performance, as shown in Table~\ref{tab:robotwin_main}. Firstly, \modelname-GD surpasses all existing VLA models with only 0.2B parameters, achieving a 90.8\% success rate under the randomization setting. Furthermore, \modelname-QW attains a higher success rate of 92.3\%. These results demonstrate the superior precision and stability of \modelname in multi-task execution.

\textbf{\calmfont{LIBERO and LIBERO-Plus Benchmark.}}
LIBERO~\cite{liu2023libero} is a widely used benchmark for evaluating robotic policies, consisting of four task suites: LIBERO-Spatial, LIBERO-Object, LIBERO-Goal, and LIBERO-Long. 
Each suite includes 10 tasks, with 50 human-teleoperated demonstrations per task. 
Following the protocol in OpenVLA~\cite{kim2024openvla}, we replay the official demonstrations in simulation and discard any unsuccessful trajectories before training. For evaluation, we run 50 independent trials per task, yielding 500 rollouts per suite.
A key limitation of the original LIBERO benchmark is its limited environmental diversity: the training and testing conditions are nearly identical. As a result, the extremely high reported success rates (often exceeding 95\%) may reflect overfitting rather than true robustness. To address this issue, LIBERO-Plus~\cite{fei2025libero} introduces multi-dimensional perturbations for rigorous generalization assessment across seven axes: object arrangement, camera viewpoints, robot initial states, language instructions, lighting, background textures, and sensor noise. The benchmark comprises 10,030 diverse evaluation rollouts in total.
Following the standardized zero-shot protocol in LIBERO-Plus, we directly evaluate our policy trained on the original LIBERO dataset, without any additional fine-tuning.

As detailed in Table~\ref{tab:exp_libero}, our approach demonstrates great performance across both standard and robust evaluation protocols. On the standard LIBERO benchmark, \modelname-QW achieves a 97.4\% success rate, performing on par with leading methods like X-VLA (98.1\%) and OpenVLA-OFT  (97.1\%).
Crucially, in the zero-shot LIBERO-Plus benchmark which designed to rigorously assess OOD generalization, our \modelname-GD sets a new state-of-the-art with an average score of 74.0\%. This significantly outperforms the previous best method, OpenVLA-OFT (69.6\%), and the X-VLA (69.7\%), confirming that our policy learns robust, transferable manipulation primitives rather than merely memorizing training distributions.

%

%

\begin{table}[t]
\centering
\caption{
Evaluation on LIBERO and LIBERO-Plus benchmarks.
For LIBERO-Plus, we evaluate the model trained on LIBERO in a zero-shot manner, without further fine-tuning, across seven distinct distribution shifts: Camera, Robot, Language, Light, Background, Noise, and Layout.
}
\label{tab:exp_libero}
\resizebox{\textwidth}{!}
{
\begin{tabular}{l|c|ccccc|cccccccc}
\toprule
\multirow{2}{*}{\textbf{Methods}} & \multirow{2}{*}{\textbf{Size}} & \multicolumn{5}{c|}{\textbf{LIBERO}} & \multicolumn{8}{c}{\textbf{LIBERO-Plus}}  \\
 & & Spatial & Object & Goal & Long & Avg & Cam & Robot & Lang & Light & BG & Noise & Layout & Avg \\
\midrule
MemoryVLA~\cite{shi2025memoryvla} & 7B & 98.4 & 98.4 & 96.4 & 93.4 & 96.7 & - & - & - & - & - & - & - & -  \\
Octo~\cite{team2024octo} & 0.1B & 78.9 & 85.7 & 84.6 & 51.1 & 75.1 & - & - & - & - & - & - & - & -  \\
FLOWER~\cite{reuss2025flower} & 1B & 97.1 & 96.7 & 95.6 & 93.5 & 95.7 & - & - & - & - & - & - & - & - \\
SmolVLA~\cite{shukor2025smolvla} & 2B & 93.0 & 94.0 & 91.0 & 77.0 & 88.8 & - & - & - & - & - & - & - & - \\
GR00T-N1~\cite{bjorck2025gr00t} & 3B & 94.4 & 97.6 & 93.0 & 90.6 & 93.9 & - & - & - & - & - & - & - & - \\
OpenVLA~\cite{kim2024openvla} & 7B & 84.7 & 88.4 & 79.2 & 53.7 & 76.5 & 0.8 & 3.5 & 23.0 & 8.1 & 34.8 & 15.2 & 28.5 & 15.6   \\
OpenVLA-OFT~\cite{kim2025fine} & 7B & 97.6 & 98.4 & 97.9 & 94.5 & 97.1 & 56.4 & 31.9 & \textbf{79.5} & 88.7 & 93.3 & 75.8 & 74.2 &  69.6  \\
WorldVLA~\cite{cen2025worldvla} & 1B & 85.6 & 89.0 & 82.6 & 59.0 & 79.1 & 0.1 & 27.9 & 41.6 & 43.7 & 17.1 & 10.9 & 38.0 & 25.0  \\
UniVLA~\cite{bu2025univla} & 7B & 96.5 & 96.8 & 95.6 & 92.0 & 95.2 & 1.8 &  46.2 & 69.6 & 69.0 & 81.0 & 21.2 & 31.9 & 42.9  \\
$\pi_0$~\cite{pi0} & 3B & 96.8 & 98.8 & 95.8 & 85.2 & 94.1 & 13.8 & 6.0 & 58.8 & 85.0 & 81.4 & \textbf{79.0} & 68.9 & 53.6   \\
$\pi_0$+FAST~\cite{fast} & 3B & 96.4 & 96.8 & 88.6 & 60.2 & 85.5 & 65.1 & 21.6 & 61.0 & 73.2 & 73.2 & 74.4 & 68.8 & 61.6  \\
$\pi_{0.5}$~\cite{intelligence2025pi0} & 3B  & 98.8 & 98.2 & 98.0 & 92.4 & 96.9 & - & - & - & - & - & - & - & -  \\
X-VLA~\cite{zheng2025x} & 0.9B  & 98.2 & 98.6 & 97.8 & 97.6 & 98.1 & 22.2 & \textbf{87.8} & 73.1 & 88.2 & \textbf{95.3} & 61.8 & 70.7 & 69.7  \\
\midrule
  {\modelname-GD} & {0.2B} & 97.8 & 98.2 & 95.2 & 95.6 & 96.7 & 65.5 & 58.2 & 78.7 & 88.1 & 90.3 & 66.9 & \textbf{79.5} & \textbf{74.0}  \\
  {\modelname-QW} & {1.1B} & 97.2 & 99.6 & 97.6 & 95.2 & 97.4 & \textbf{66.3} & 49.0 & 65.9 & \textbf{94.9} & 93.3 & 73.1 & 78.2 & 72.6  \\
\bottomrule
\end{tabular}
}
\end{table}

%

\textbf{\calmfont{GenieSim 2.2 Benchmark (Agibot Challenge 2025).}}
While the benchmarks mentioned above focus on fixed-base tabletop manipulation, we extend our evaluation to the GenieSim Benchmark \cite{yin2026geniesim30}. This platform utilizes the Agibot G1, a humanoid robot featuring a wheeled mobile base and a vertically adjustable torso. Specifically, we adopt GenieSim v2.2, the version served as the basis for the Agibot Challenge 2025~\cite{contributors2025agibotdigitalworld}. This benchmark comprises 10 diverse and challenging manipulation tasks.
Each task is evaluated over 25 trials, totaling 250 evaluation episodes.
We traine a multi-task model to evaluate all tasks and tune the effective steps used in the action chunk for each task individually. As shown in Table~\ref{tab:exp_geniesim}, \modelname-QW achieves an overall score of 4.685, surpassing X-VLA’s score of 4.541. This result demonstrates that our model can deliver strong performance  on humanoid (upper-body) embodiments.

\begin{table}[h]
    \centering
    \caption{
    Evaluation on the GenieSim 2.2 benchmark. 
    Baselines are sourced from the official leaderboard, limited to methods with associated papers. Values denote progress scores.
    }
    \label{tab:main_results}
    { 
    \begin{tabular}{lcccc}
        \toprule
        Task & RDT~\cite{rdt} & UniVLA~\cite{bu2025univla} & X-VLA~\cite{zheng2025x} & \modelname-QW \\
        \midrule
        Clear the countertop waste        & 0.296 & 0.097 & \textbf{0.622} & 0.398 \\
        Open drawer and store items       & 0.000 & 0.020 & \textbf{0.400} & 0.152 \\
        Heat the food in the microwave    & 0.133 & 0.033 & \textbf{0.524} & 0.518 \\
        Pack moving objects from conveyor & 0.000 & \textbf{0.350} & 0.280 & \textbf{0.350} \\
        Pickup items from the freezer     & \textbf{0.480} & 0.260 & 0.424 & 0.472\\
        Restock supermarket items         & 0.250 & 0.400 & 0.610 & \textbf{0.820} \\
        Pack in the supermarket           & 0.825 & \textbf{1.000} & 0.940 & 0.940 \\
        Make a sandwich                   & 0.000 & 0.080 & 0.143 & \textbf{0.145} \\
        Clear table in the restaurant     & 0.250 & 0.375 & 0.310 & \textbf{0.570} \\
        Stamp the seal                    & 0.200 & 0.180 & 0.288 & \textbf{0.320} \\
        \midrule
        Total Score                       & 2.434 & 2.795 & 4.541 & \textbf{4.685} \\
        \bottomrule
    \end{tabular}
    }
\label{tab:exp_geniesim}
\end{table}

\subsection{Ablation and Analysis}
\textbf{\calmfont{Co-Training with Grasp Anything.}}
To validate whether co-training with auxiliary task data can mitigate insufficient post-training data, we conduct an ablation study where we jointly train on the Grasp Anything task and seven basic tasks. The Grasp Anything task, with its diverse objects (varying shapes, materials, poses) and focus on robust grasping, is chosen specifically to enhance the generalization of the grasp skill—a foundational capability critical for nearly all basic tasks. As shown in Table~\ref{tab:cotraining}, augmenting training with Grasp Anything data yields significant performance gains across the seven basic tasks, with average success rates increasing markedly compared to training on basic tasks alone. These results highlight that leveraging data from a skill-focused auxiliary task (like Grasp Anything) is a viable approach to strengthen core competencies and improve generalization on simpler tasks.

\begin{table}[]
    \centering
    \caption{Evaluation of the effectiveness of co-training fundamental tasks with ``Grasp Anything'' data. Each cell follows the format: ``progress score / success rate".}
    \begin{tabular}{c|cccc}
    \toprule
        Co-Training  & Fold towel & Place empty cup & Place shoe & Stack blocks three \\
    \midrule
         &  84.74 / 84.21 & 78.50 / \textbf{70.00} & 93.23 / \textbf{93.55} & 83.33 / 46.67 \\
        \checkmark &\textbf{88.95} / \textbf{89.47} & \textbf{82.00} / \textbf{70.00} & \textbf{95.16} / \textbf{93.55} & \textbf{91.11} / \textbf{66.67} \\
    \midrule
    Co-Training  &  Stack bowls three & Put bottles basket & Place to slot & Average\\
    \midrule
    & \textbf{100.00} / \textbf{100.00} & 95.24 / \textbf{85.71} & 62.22 / 26.67 & 85.32  / 72.40 \\
    \checkmark & 99.56 / 93.33 & \textbf{96.67} / 78.57 & \textbf{70.89} / \textbf{33.33} & \textbf{89.19} \  \textbf{75.00} \\
    \bottomrule
    \end{tabular}
    \label{tab:cotraining}
\end{table}

\begin{figure}[t]
  \centering

  \begin{minipage}[t]{0.49\textwidth}
    \centering
    \includegraphics[width=\linewidth]{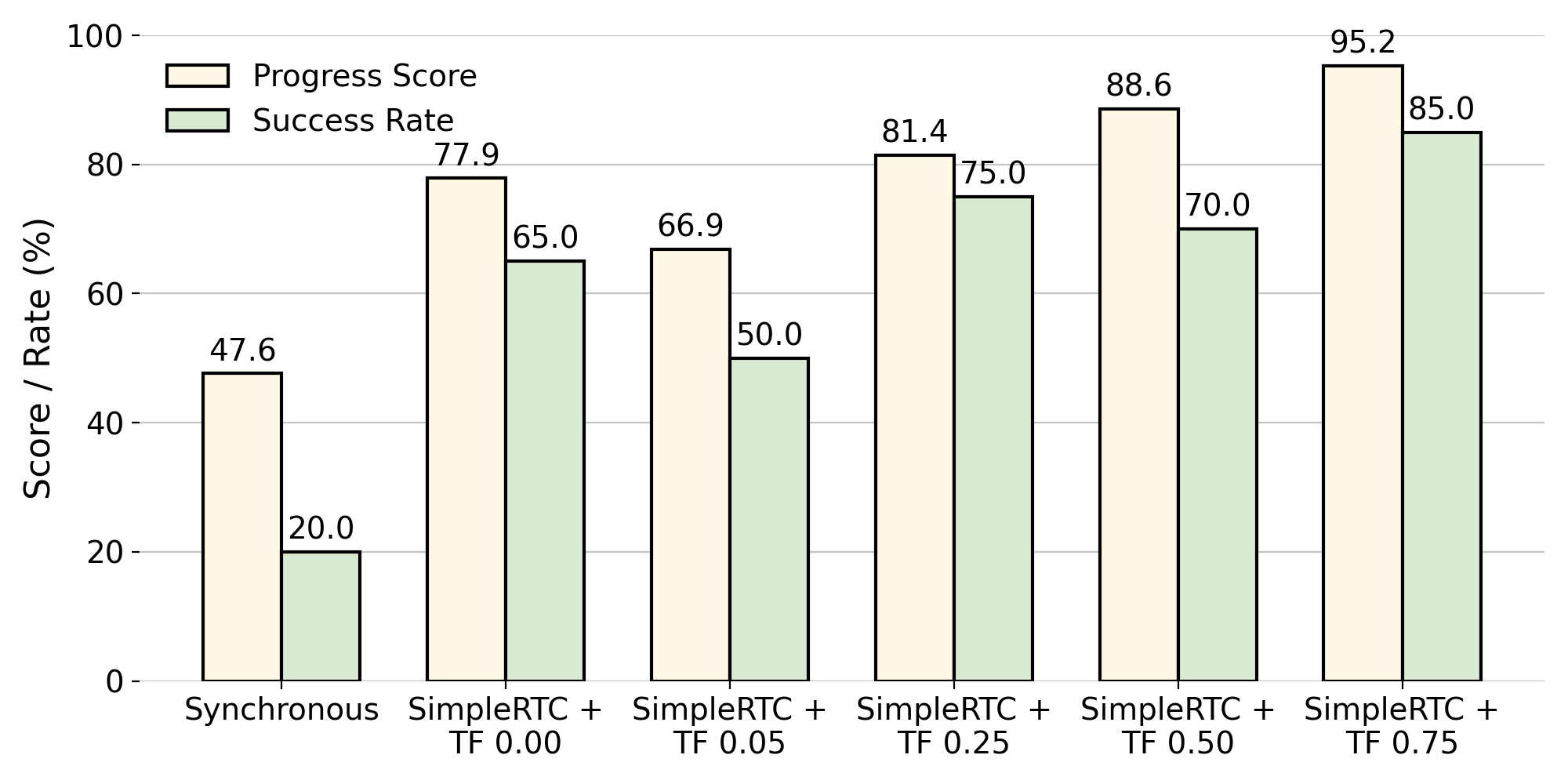}
    \label{fig:rtc_left}
  \end{minipage}
  \hfill
  \begin{minipage}[t]{0.49\textwidth}
    \centering
    \includegraphics[width=\linewidth]{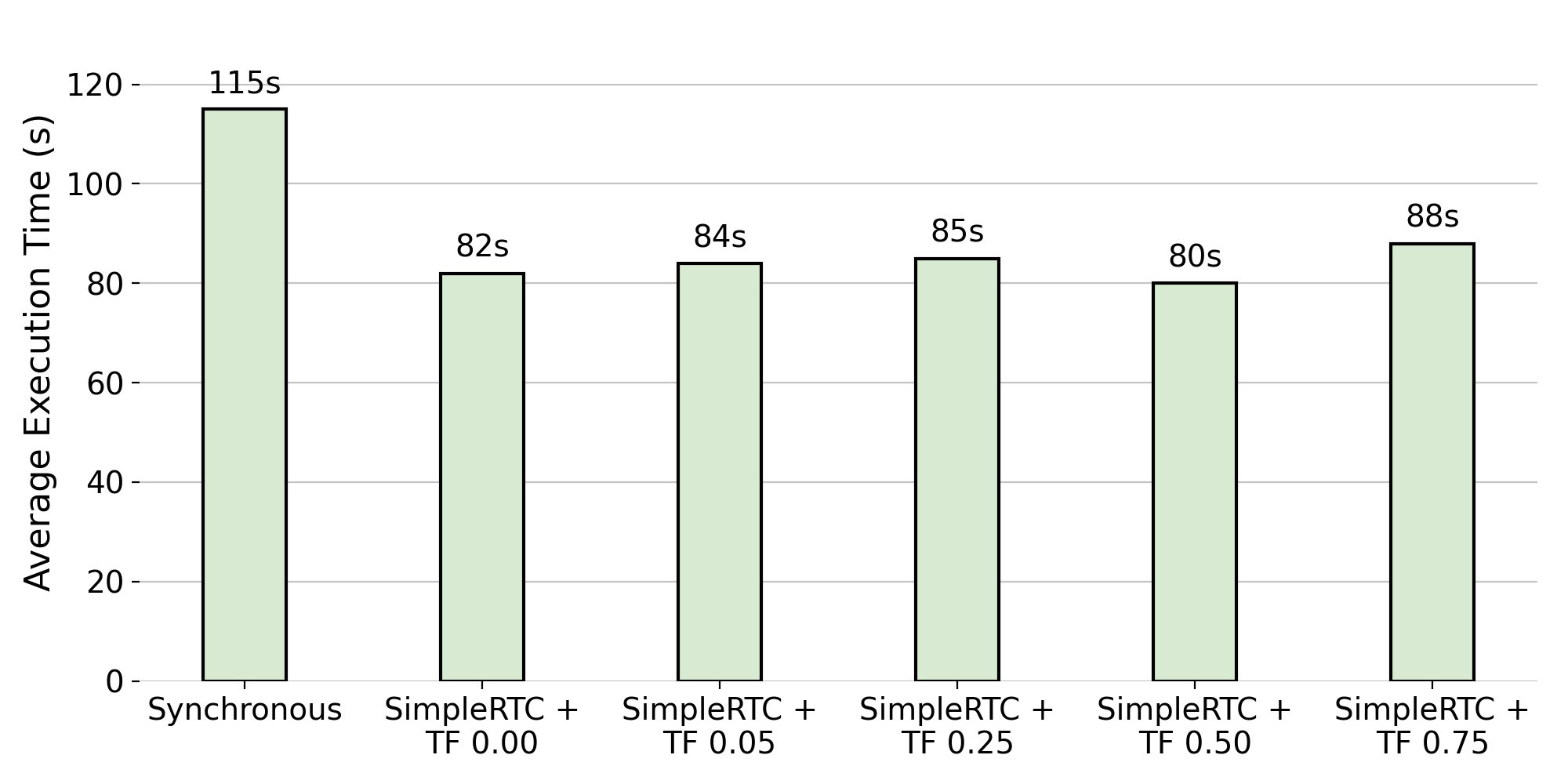}
    \label{fig:rtc_right}
  \end{minipage}

\vspace{-4mm}
  \caption{
Analysis of the proposed smooth asynchronous inference strategy, including SimpleRTC and Teacher Forcing (TF) training.
Experiments are conducted on the cloth folding task.
We benchmark a synchronous baseline against five models utilizing SimpleRTC trained with different teacher-forcing rates. (Left) Comparison of success rates and scores. (Right) Average inference time for successful episodes.
  }
  \label{fig:rtc_combined}
\end{figure}

\textbf{\calmfont{How to Achieve Smooth Asynchronous Inference.}}
%
%
We evaluated our SimpleRTC module and Teacher Forcing (TF) training strategy through an ablation study on the cloth folding task. Because cloth folding requires highly smooth motion, it serves as a strong benchmark for testing asynchronous inference.
Our experimental setup compared a synchronous inference baseline against five models utilizing SimpleRTC, each trained with varying teacher-forcing ratios ranging from 0\% (without TF) to 75\%. 
As illustrated in Fig.~\ref{fig:rtc_combined}, SimpleRTC-based asynchronous inference significantly outperforms the synchronous baseline in both success rates and progress scores. 
%
Furthermore, while higher teacher-forcing ratios yield substantial gains for dexterous tasks, very low ratios (e.g., 5\%) prove detrimental to performance.
In Table~\ref{tab:real_world}, we adopted a unified ratio of 25\% to balance performance across diverse tasks. 
Regarding efficiency, Fig.~\ref{fig:rtc_combined} (b) demonstrates that while SimpleRTC significantly reduces task completion time compared to the synchronous baseline, varying the teacher-forcing rate has a negligible impact on efficiency. In conclusion, our proposed inference-time and training-time strategies exhibit strong complementarity, effectively eliminating inference pauses and achieving smooth control for complex dynamic tasks.

\section{Related Work}
\label{supp_sec:related_work}

\textbf{\calmfont{Action Model for Manipulation.}}
End-to-end robot manipulation model have advanced rapidly, with progress achieved from multiple perspectives. Some approaches focus on modeling the probability distribution of actions, proposing methods such as implicit behavioral cloning~\cite{IBC}, gaussian mixture models~\cite{BeT}, and diffusion~\cite{dp}. Others explore designing more suitable network architectures to improve manipulation performance~\cite{dp3,peract,gervet2023act3d,kstar,3ddiffuseract,goyal2023rvt,goyal2024rvt-2}.
Over the past two years, spurred by the rapid advancement of VLMs~\cite{chen2023pali,llava,beyer2024paligemma,wang2024qwen2}, an increasing number of studies have integrated VLMs with action models to build end-to-end vision-language-action systems. These methods are typically trained on large-scale action datasets, aiming to endow models with strong visual generalization and instruction following capabilities. OpenVLA~\cite{kim2024openvla} augments word vocabulary with discrete action tokens and generates actions autoregressively via a language model. $\pi_0$~\cite{pi0,intelligence2025pi0} attaches an additional action expert to the VLM and leverages flow matching to model the action distribution. FAST~\cite{fast} introduces the discrete cosine transform to encode actions for higher training efficiency. $\pi_{0.6}$~\cite{intelligence2025pi} combines offline reinforcement learning and decision transformer to boost performance. SpatialVLA~\cite{sptialVLA} enhances VLMs with 3D positional embeddings to improve spatial reasoning, while X-VLA~\cite{zheng2025x} investigates efficient heterogeneous data mixing strategies. LingBot-VLA~\cite{wu2026pragmatic} utilizes larger-scale real-world action data to enhance model generalization.
Beyond those VLA models extended from VLMs, several large action models aim to improve performance by incorporating video generation. GR-1~\cite{gr1}, Motus~\cite{bi2025motus}, and LingBot-VA~\cite{lingbotva} integrate video generation tasks into the training process.
In this work, we propose \modelname, a novel model that seeks to push the performance boundaries of existing action models through superior architectural design, training methodology, and data acquisition strategy.

\textbf{\calmfont{Dataset and Data Strategy.}}
Existing data collection strategies for mitigating distribution shift generally follow three main paradigms. 
First, \textit{adversarial collection methods}, such as ADC \cite{huang2025adversarial} and MOVE \cite{wang2025move}, utilize manual perturbations to enhance model robustness. However, these approaches are often labor-intensive and susceptible to causal misalignment, particularly when predicted action chunks fail to account for stochastic external interventions. 
Second, \textit{state-space augmentation strategies}, exemplified by HDSpace \cite{yang2025bootstrapping}, focus on systematically expanding the state space through hierarchical sampling. Despite improving coverage, such methods are frequently limited to initial joint states and trajectory-level diversity, often overlooking critical variations in visual conditions and object states. 
Moreover, they typically treat all initial states with uniform importance, ignoring the disproportionate impact that specific critical states may have on overall policy performance. 
Finally, distinct from these offline approaches, \textit{interactive collection frameworks} like DAgger \cite{ross2011reduction} and Genie Centurion \cite{wang2025genie} employ human-in-the-loop corrections to rectify policy drift in real-time. Nevertheless, these methods suffer from diminishing sampling efficiency: as the policy improves, failure cases become increasingly rare, rendering large-scale data collection progressively impractical.



\textbf{\calmfont{Robotic Infrastructure.}}
Recent advancements in robotic infrastructure have centered on democratizing access to embodied AI. Platforms like LeRobot \cite{cadene2024lerobot}, built upon the Hugging Face ecosystem, represent a significant leap forward by standardizing multimodal data formats and providing accessible PyTorch implementations of policies such as ACT and Diffusion Policy. By integrating hardware interfaces with cloud-hosted datasets, LeRobot effectively lowers the barrier to entry, enabling rapid prototyping and cross-institutional data sharing \cite{o2024open}.
However, while such platforms streamline development accessibility, they often face bottlenecks in high-throughput data handling and deployment reproducibility. 
Addressing these limitations, our framework RoboOrchard prioritizes efficiency by adopting the \textit{MCAP} format for robust logging and \textit{Apache Arrow} for zero-copy training data loading, significantly accelerating I/O compared to standard serialization. 
Furthermore, to bridge the fragmentation between simulation and reality often found in existing codebases, we implement a \textit{Unified Environment Interface} alongside a rigorous \textit{Model Artifact} system; unlike basic checkpoints, these artifacts encapsulate the full inference context to ensure self-contained deployment. 
Finally, moving beyond CLI-based workflows, we introduce a no-code \textit{Companion App}, simplifying the entire lifecycle from data acquisition to model deployment.


\section{Conclusion}
\label{sec:conclusion}

In this work, we presented \modelname, a holistic Vision-Language-Action (VLA) framework designed to bridge the gap between foundation model research and reliable real-world deployment. By explicitly injecting embodiment priors, such as kinematic chains and multi-view camera parameters, our architecture enables robust 3D spatial reasoning and cross-embodiment generalization. 
Through a rigorous ``pre-train then post-train'' paradigm supported by the open-source \textit{RoboOrchard} infrastructure, \modelname achieves state-of-the-art performance across simulation benchmarks (e.g., RoboTwin 2.0, LIBERO, LIBERO-Plus, GenieSim) and challenging real-world manipulation tasks. 
Finally, by open-sourcing the entire ecosystem, including pre-trained foundations, post-trained checkpoints, and the software stack, we provide the community with a complete, reproducible path toward high-performance generalist robotic agents.

\section{Future Work}
\label{sec:future_work}

We are actively developing the next iteration of the HoloBrain ecosystem. The primary objective of our upcoming release is to enhance the VLA policy's task success rate, instruction adherence, and generalization capabilities, all while minimizing data acquisition costs. Specifically, we will focus on three key directions:

\begin{itemize} [leftmargin=*]

\item \textbf{Off-Policy Reinforcement Learning:} While  \modelname focuses on imitation learning, our next version will integrate off-policy reinforcement learning algorithms. This expansion aims to maximize the utility of diverse data sources, including expert demonstrations, policy rollouts, and human interventions. To support this, we plan to architecturally integrate  a Value Model within the VLA framework.

\item \textbf{Rigorous Instruction Following:} 
%
We observe that precise instruction following is still an under-evaluated ability in current VLA research, especially when instructions are easy to confuse. Future work will focus on building clearer and more thorough benchmarks to assess this ability, along with improving the model’s reliability in carrying out easily confusable instructions.

\item \textbf{Advanced Co-training Strategies:} 
%
%
Our preliminary results show that mixing data from different tasks during post-training leads to notable synergistic gains. This suggests that co-training can help break down complex capabilities across diverse datasets. We will further explore these strategies to improve generalizability and support efficient few-shot transfer to new robot embodiments.

\end{itemize}





\newlength{\vertsep}
\setlength{\vertsep}{.085in}
\newlength{\imsize}
\setlength{\imsize}{.365\textwidth}

\clearpage

\bibliography{paper}
\bibliographystyle{unsrt}

\clearpage

\appendix
\renewcommand{\thefigure}{A\arabic{figure}}
\renewcommand{\thetable}{A\arabic{table}}
\setcounter{figure}{0}
\setcounter{table}{0}

\section{\modelname Implementation Details}
\label{supp_sec:model}

\begin{figure}[h]
    \centering
    \includegraphics[width=0.8\linewidth]{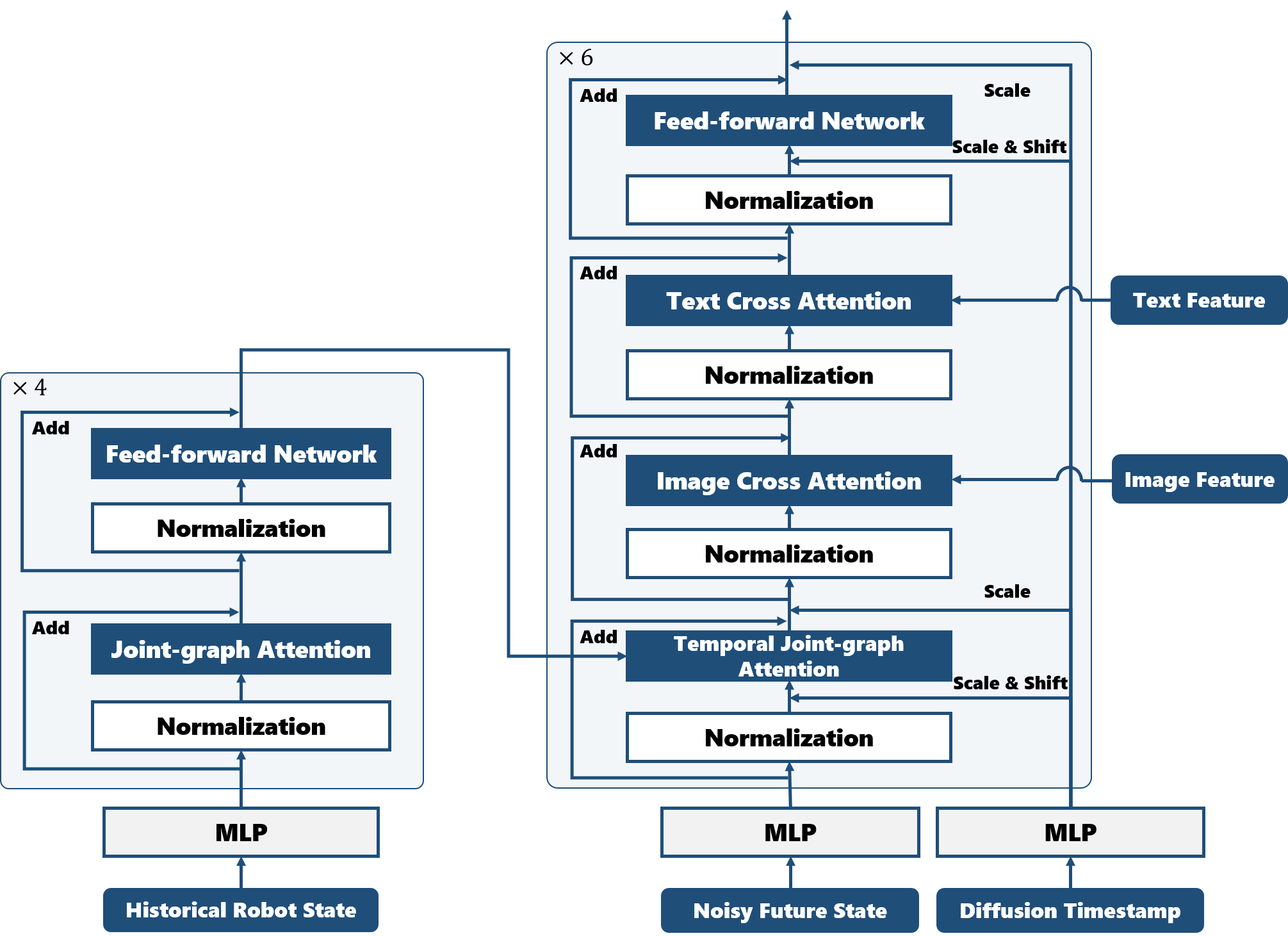}
    \caption{Detailed architecture of Action Expert in \modelname}
    \label{fig:model_details}
\end{figure}

We employ two versions of vision-language models as our base models: GroundingDINO Tiny and Qwen-2.5-VL-3B, which are used to construct HoloBrain-0-GD and HoloBrain-0-QW, respectively.
For GroundingDINO Tiny, all architectural components and parameters are retained, with the BERT~\cite{devlin2019bert} module frozen during training. For Qwen-2.5-VL, the language model is pruned by retaining only the first layer, with the vision encoder frozen.
Beyond the base models, we additionally introduce two randomly initialized modules: the spatial enhancer and the action expert. The spatial enhancer adopts a Swin Transformer~\cite{swin} as its backbone to encode the depth map, and fuses the encoded depth features with the 2D image features.
The action expert comprises two transformers, one for encoding the current robot state and another for decoding the future robot state; the detailed architecture is illustrated in the Fig.~\ref{fig:model_details}.
The hyperparameters involved in the model and training are listed in Tab.~\ref{tab:hyperparameters}.
Notably, under the image resolution settings in Tab.~\ref{tab:hyperparameters}, \modelname-QW and \modelname-GD can be configured with single-GPU batch sizes of 32 and 8, respectively, requiring only approximately 30 GB of GPU memory without any optimization techniques.
This implies that the \modelname can be trained on a wider range of GPUs.
Note that \modelname-GD requires more GPU memory because we unfroze its vision encoder.

\begin{table}[]
    \centering
    \caption{Hyperparameter settings of \modelname. Parameters not explicitly specified for post-training remain the same as in pretraining, such as learning rate and weight decay.}

    \begin{tabular}{l|c|c}
    \toprule
         & \textbf{Configuration} & \textbf{Value} \\
    \midrule
    \multirow{9}{*}{\textbf{Model}}& action expert embedding channel & 256 \\
    & action expert FFN channel & 2048 \\
    & attention heads & 8 \\
    & action decoder layers & 6\\
    & state encoder layers & 4 \\
    & prediction steps & 64 \\
    & historical steps & 1 \\
    & normalization & RMSNorm \\
    & activation & SiLU \\
    \midrule
        \multirow{10}{*}{\textbf{Pretraining}}& batch size for \modelname -QW  & 2048 \\
        & batch size for \modelname -GD  & 512 \\
        & learning rate & $10^{-4}$ \\
        & optimizer & AdamW \\
        & weight decay & 0.0005 \\
        & training steps & $2\times 10^5$ \\
        & teacher forcing $N_{\text{prefix}}$ & 16 \\
        & smooth L1 beta & 0.04 \\
        & candidate trajectories $N$ & 4 \\
        & warmup steps & 500 \\
        & image size for \modelname -QW & $308\times 252$\\
        & image size for \modelname -GD & $320\times 256$\\
    \midrule
        \multirow{7}{*}{\textbf{Post-training}} & batch size for \modelname -QW & 256 \\
        & batch size for \modelname -GD & 128 \\
        &training steps for 7 real-world tasks & $1.2\times 10^5$ \\
        &training steps for 3 long-term tasks & $2\times 10^5$ \\
        &training steps for RoboTwin & $2\times 10^5$ \\
        &training steps for LIBERO & $2\times 10^5$ \\
        &training steps for GenieSim &  $2\times 10^5$  \\
        &lr scheduler & $\times 0.1$ for the last 10\% steps \\
    \midrule
    \multirow{5}{*}{\textbf{Diffusion}} & training noise scheduler & DDPM \\
    & inference noise scheduler & DPMSolver++\\
    &training diffusion steps & 1000\\
    &inference diffusion steps & 10\\
    & beta schedule & squaredcos cap v2 \\
    \bottomrule
    \end{tabular}
    \label{tab:hyperparameters}
\end{table}

\newpage

\section{Real-World Experimental Details}
\label{supp_sec:real}

\begin{figure}[h]
    \centering
    \includegraphics[width=0.8\linewidth]{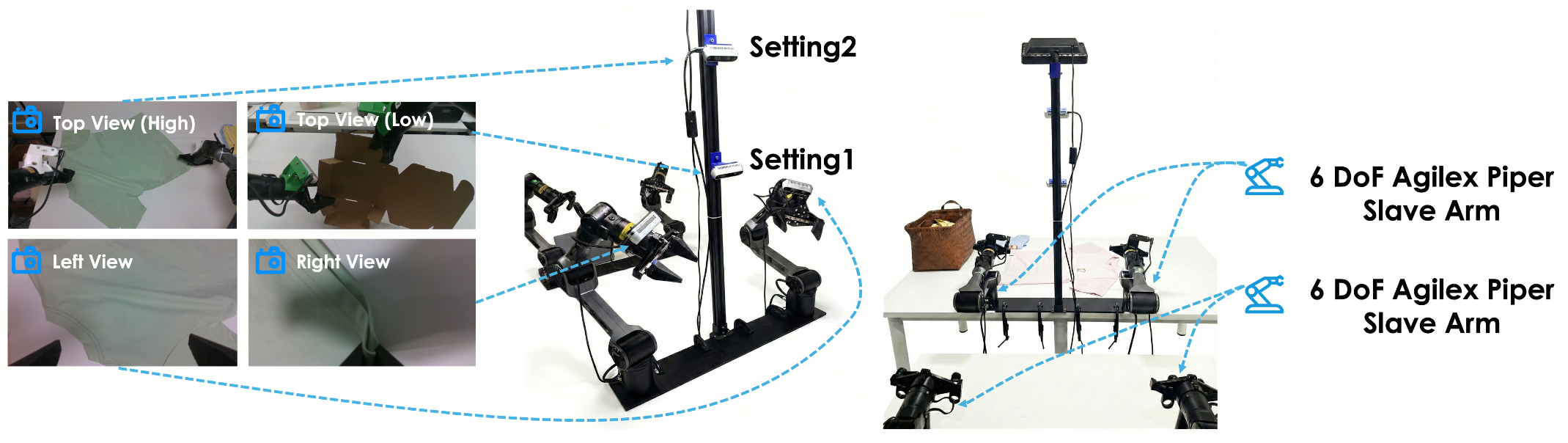}
    \caption{Illustration of the real-world experimental setup. Two global camera configurations are implemented to accommodate the specific field-of-view requirements of different tasks.}
    \label{fig:robot_setup}
\end{figure}

We conduct real-world experiments using a dual-arm AgileX Piper robot equipped with three Intel RealSense D435 RGB-D cameras. Data collection is performed via homologous teleoperation, following a paradigm similar to ALOHA~\cite{aloha}. The hardware setup is illustrated in Fig.~\ref{fig:robot_setup}. All tasks involve fixed-base tabletop manipulation. To accommodate the extensive workspace required for the cloth folding task, we employ two distinct global camera configurations: Setting 1 (the default for standard tasks) and Setting 2 (optimized for cloth folding). Specifically, Setting 2 elevates the camera by 25 cm relative to Setting 1, with a minor adjustment to the pitch angle.

\begin{figure}[h]
    \centering
    \includegraphics[width=0.8\linewidth]{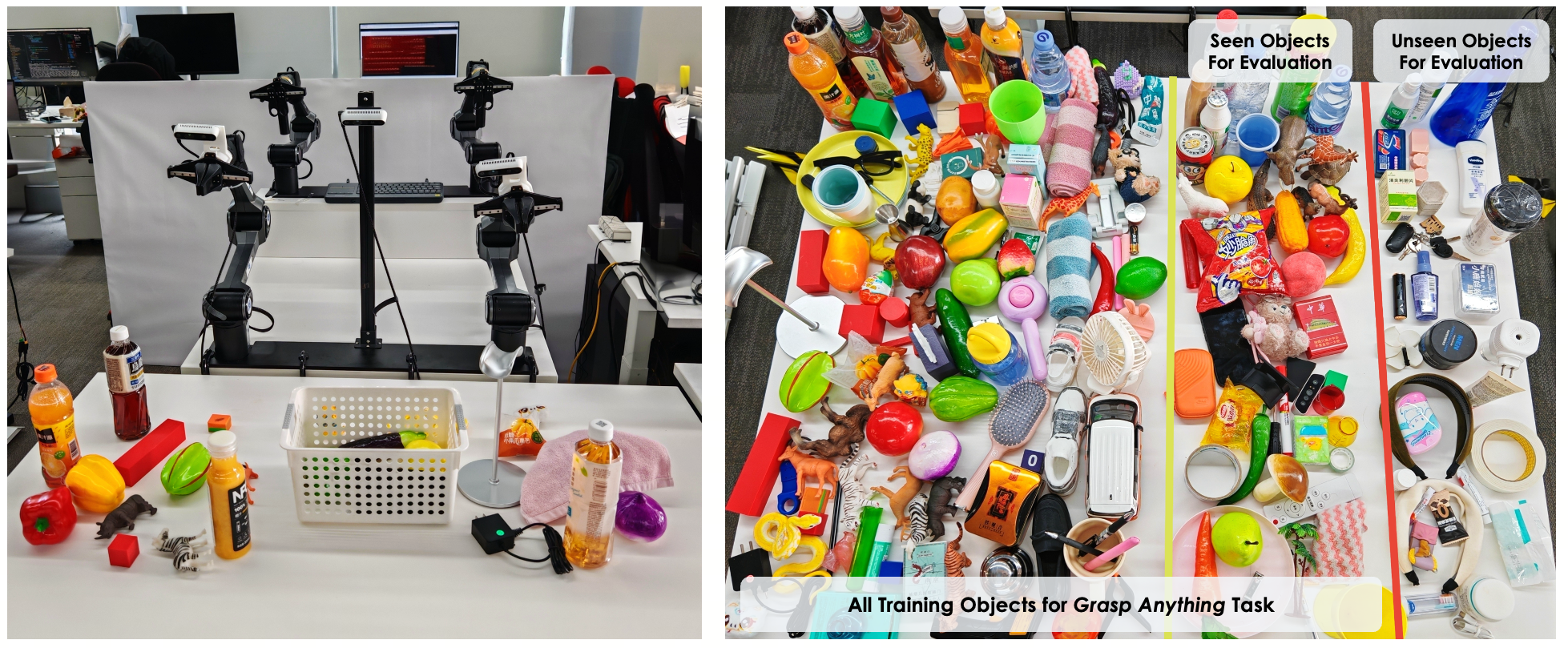}
    
    \caption{
    Visualization of the "Grasp Anything" task. (Left) The scenario during data collection. (Right) All training objects, as well as the seen and unseen objects used during evaluation.
    }
    \label{fig:vis_grasp_anything}
\end{figure}

For the ``Grasp Anything" task, we collected a large-scale pick-and-place dataset involving over 200 distinct objects. The data acquisition process was highly efficient, characterized by continuous grasping sequences of dozens of items per episode. As illustrated in Fig.~\ref{fig:vis_grasp_anything} (a), the object set spans a broad spectrum of categories, including rigid geometries, irregular shapes, deformable materials, and semi-transparent items. During evaluation (Fig.~\ref{fig:vis_grasp_anything} (b)), we tested the model on a mix of objects seen during training and a suite of novel, unseen objects. The results demonstrate comparable success rates across both groups, suggesting that the model has acquired robust, object-agnostic grasping capabilities and exhibits strong generalization.


\begin{figure}[tb]
     \centering
     \includegraphics[width=0.9\linewidth]{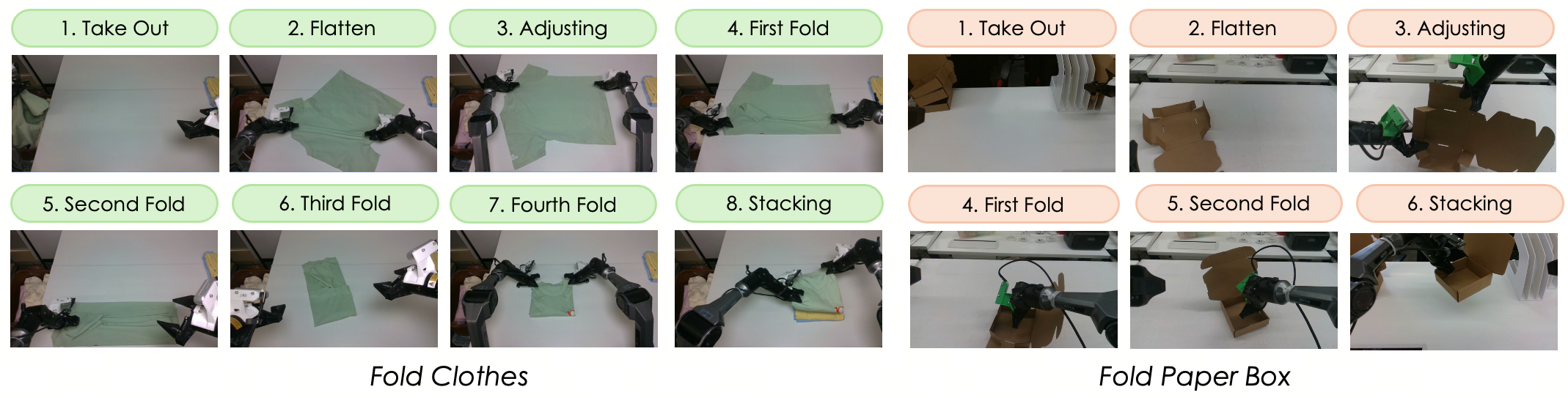}
     
     \caption{
     Illustration of the Standard Operating Procedures (SOPs) for two long-horizon dexterous tasks. 
     The cloth folding task consists of 8 distinct stages, while the paper box folding task involves 6 stages.
     }
     \label{fig:task_sop}
 \end{figure}

In Table~\ref{tab:real_sop}, we detail the language instructions, data collection protocols (SOPs), and definitions of progress scores for all 10 tasks. 
During data collection, while adhering to these protocols, we maximized the spatial diversity of initial object placements within the available workspace.
Meanwhile, for evaluation, we established a standardized object placement guide. We strictly followed these pre-defined configurations during testing to ensure consistency and comparability across trials. Additionally, Fig.~\ref{fig:task_sop} illustrates the detailed stage-wise workflow for the two long-horizon dexterous tasks.

\begin{table}[h]
\centering
\small
\begin{tabularx}{\textwidth}{l p{2.5cm} X X}
\toprule
\textbf{Task Name} & \textbf{Instruction} & \textbf{Collection SOP} & \textbf{Progress Score Definition} \\ \midrule

Place Empty Cup & Place an empty cup on a mat. & Use the arm corresponding to the cup's side (L/R). Top-down grip on the wall. Place on mat and retract. & \textbf{Total 1.0:} Grip (0.5), On mat (0.2), Perfectly placed (0.3). \\ \midrule

Place Shoe & Place a shoe on a mat using one arm. & Use same-side arm. Top-down grip at the middle. Place on mat and retract. & \textbf{Total 1.0:} Lift (0.4), Partial on mat (0.2), Fully on mat (0.2), Direction (0.2). \\ \midrule

Stack Bowls Three & Stack three bowls. & L-arm for left, R-arm for right, nearest arm for center. Grip sides (L/R arm accordingly). & \textbf{Total 1.5:} Pick (0.25), Misaligned stack (+0.15), Perfect stack (+0.1). Max 3 attempts/bowl. \\ \midrule

Stack Blocks Three & Stack cubes in RGB order. & Side-specific arm usage. Sequence: Red (bottom) $\rightarrow$ Green $\rightarrow$ Blue (top). Top-down grip. & \textbf{Total 1.5:} For 3 blocks, Pick (0.25) and Place (0.25) per block. \\ \midrule

Put Bottles Dustbin & Put 3 bottles into the bin. & L-arm: bottles 1 \& 2. R-arm: hand bottle 3 to L-arm. Horizontal grip. Retract arms. & \textbf{Total 1.5:} For 3 bottles, Pick (0.3) and Place (0.2) per bottle. \\ \midrule

Place to Slot & Insert blocks into matching slots. & L-arm: Square. R-arm: Cylinder/T-shape. Top-down grip. Must fully enter the slot. & \textbf{Total 1.5:} Pick (0.25), Within 2cm (+0.15), Perfect placement (+0.1). Max 3 attempts/block. \\ \midrule

Two Fold Towel & Fold a towel twice. & Center towel. Dual arms pick bottom corners and fold up. R-arm folds right edge to the left. & \textbf{Total 1.0:} Pick corner (0.1), 1st fold (0.3), Pick right (0.1), 2nd fold (0.3), Quality (0.2). \\ \midrule

Fold Clothes & Fold a T-shirt four times. & Orient shirt. 1st/2nd folds on sides; 3rd/4th folds after adjustment. Stack or store in top-left. & \textbf{Total 21:} Orientation (1), Folds 1-4 (3, 4, 4, 4), Stacking/Storage (5). \\ \midrule

Grasp Anything & Clear objects into a basket. & Move basket to center. Alternating arms pick objects and place in bin until finished. & \textbf{Total 1.0 (10 objects):} Success (0.1), Pick only (0.05). Max 3 attempts/object. \\ \midrule

Fold Paper Box & Fold a paper box. & L-arm grip, R-arm buckle. Fold sequence: 1st $\rightarrow$ 2nd $\rightarrow$ Move to top-left. & \textbf{Total 1.0:} L-grip (0.2), R-buckle (0.1), 1st fold (0.3), 2nd fold (0.3), Store (0.1). \\ \bottomrule

\end{tabularx}
\caption{Task SOP and progress score definition of 10 real-world tasks. Success criteria: For \textit{place shoe}, \textit{fold towel}, and \textit{place to slot}, a score greater than 0.9, 0.9, and 1.3, respectively, is deemed successful. For the remaining tasks, success is defined as attaining the maximum achievable score.}
\label{tab:real_sop}
\end{table}

\clearpage

 \begin{figure}[tb]
     \centering
     \includegraphics[width=0.9\linewidth]{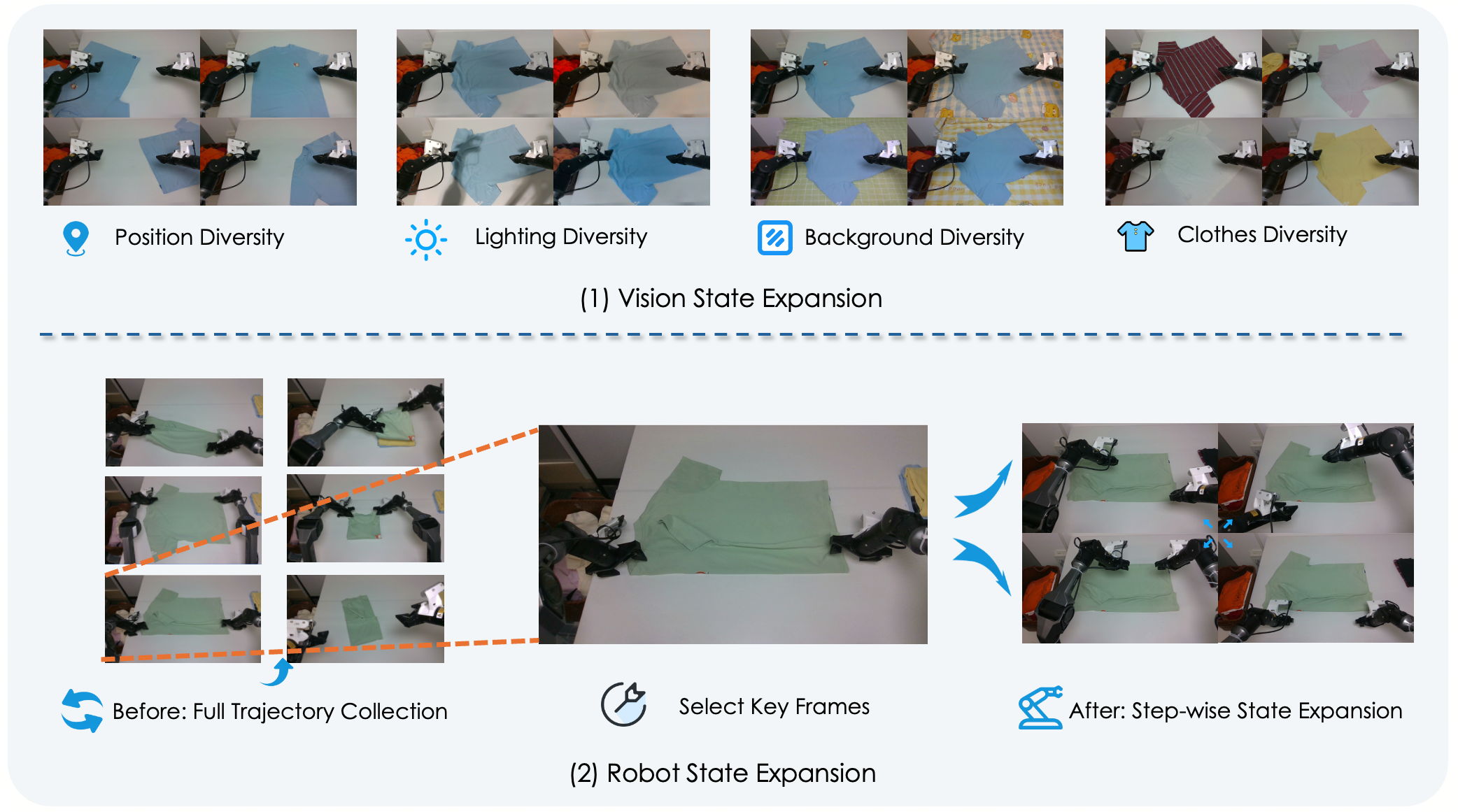}
     \caption{
     Overview of the Proactive State Expansion strategy. (1) Vision State: We actively augment the training distribution with diverse environmental factors, including variations in position, lighting, background, and object appearance. (2) Robot State: The method transitions from collecting redundant full trajectories to an efficient, step-wise expansion focused on critical key frames.
     }
     \label{fig:state_expansion}
 \end{figure}

\section{Empirical Analysis of Test-driven Performance Evolution}
\label{supp_sec:sop}

During the post-training phase, Standard Operating Procedures (SOPs) are typically established to ensure data consistency when collecting demonstrations for target tasks. However, strict adherence to SOPs introduces a trade-off: while it yields standardized datasets, it significantly reduces the information entropy of individual episodes. Consequently, models trained on such low-variance data risk overfitting to a narrow distribution, resulting in poor generalization. To address this, we implement the \textit{Iterative Test-Driven Data Strategy} (Sec.~\ref{subsec:data_strategy}). In this section, we detail the implementation of this strategy with  \textit{cloth-folding} task.


\subsection{Proactive State Expansion}

Using the cloth-folding task as a primary example, we first construct an SOP (Fig.~\ref{fig:task_sop}) based on backward induction. This method specifies the target states in advance and then identifies the key waypoints by tracing the task in reverse. During data collection, we incorporate proactive state expansion to improve state diversity. Vision-based state expansion, illustrated in Fig.~\ref{fig:state_expansion}(1), is applied during full-pipeline demonstrations by deliberately increasing visual variation. This includes modifying background environments, adjusting lighting conditions, and ensuring diverse spatial configurations of the cloth. The objective is to maximize the incremental information contributed by each episode.
For defensive robot state expansion, once a small baseline dataset has been gathered (for example, 200 full-pipeline episodes), the data-collection strategy shifts from complete demonstrations to step-wise acquisition with an emphasis on augmenting the robot state, as shown in Fig.~\ref{fig:state_expansion}(2).


Defensive robot state expansion can be implemented in various forms; typical examples include empty-grasp recovery and step-wise state expansion. Empty-grasp recovery is a widely used strategy that enables models to remediate failed manipulation attempts. Notably, as illustrated in Fig.~\ref{fig:state_expansion}(2), step-wise state expansion is achieved through keyframe selection and state-space augmentation around trajectory waypoints. Distinct from the full-episode collection approach in HDSpace~\cite{yang2025bootstrapping}, we selectively extract brief (2–3 second) recovery segments. By focusing on high-value error-correction behaviors, this approach maximizes the information density of the training data.

\subsection{Test-Driven Failure Recovery}

\begin{table}[ht!]
    \centering
    \small 
    \renewcommand{\arraystretch}{0.9} 
    
    \caption{Corrected Failure Frequency and Success Rate (20 Trials per Iteration)}
    \label{tab:folding_corrected}
    
    \begin{tabular}{@{}lllcccc@{}}
        \toprule
         & & & \multicolumn{4}{c}{\textbf{Failure Frequency}} \\ \cmidrule(l){4-7} 
        \textbf{Stage} & \textbf{ID} & \textbf{Milestone / Action Description} & \textbf{Iter 1} & \textbf{Iter 2} & \textbf{Iter 3} & \textbf{Iter 4} \\ \midrule
        \textbf{Initial} & - & Initial Setup / Score 0 & 0\% & \textbf{15\%} & 0\% & \textbf{10\%} \\ \midrule
        \textbf{Preparation} & 0 & Complete pre-folding organization (Score 1) & 0\% & 0\% & 0\% & \textbf{5\%} \\ \midrule
        \textbf{First Fold} & 1.1 & Capture the first fold point (Score 2) & \textbf{5\%} & 0\% & 0\% & 0\% \\
         & 1.2 & Complete the first flipping action (Score 3) & \textbf{5\%} & 0\% & 0\% & 0\% \\
         & 1.3 & Complete the first sleeve fold (Score 4) & 0\% & 0\% & 0\% & 0\% \\ \midrule
        \textbf{Second Fold} & 2.1 & Perform downward transition adjustment (Score 5) & 0\% & 0\% & 0\% & 0\% \\
         & 2.2 & Capture the garment corner (Score 6) & 0\% & 0\% & 0\% & 0\% \\
         & 2.3 & Complete the second flipping action (Score 7) & \textbf{35\%} & \textbf{10\%} & \textbf{10\%} & 0\% \\
         & 2.4 & Complete the second sleeve fold (Score 8) & 0\% & 0\% & 0\% & 0\% \\ \midrule
        \textbf{Third Fold} & 3.1 & Transition to target position 3 (Score 9) & \textbf{5\%} & 0\% & \textbf{10\%} & 0\% \\
         & 3.2 & Capture the garment corner (Score 10) & \textbf{5\%} & 0\% & 0\% & 0\% \\
         & 3.3 & Complete the third fold (Score 11) & 0\% & 0\% & 0\% & \textbf{5\%} \\
         & 3.4 & Complete the nudging adjustment (Score 12) & \textbf{10\%} & \textbf{5\%} & 0\% & 0\% \\ \midrule
        \textbf{Fourth Fold} & 4.1 & Transition to target position 4 (Score 13) & \textbf{10\%} & \textbf{10\%} & 0\% & 0\% \\
         & 4.2 & Capture the garment corner (Score 14) & 0\% & 0\% & 0\% & 0\% \\
         & 4.3 & Complete the fourth fold (Score 15) & 0\% & 0\% & 0\% & \textbf{5\%} \\
         & 4.4 & Complete the nudging adjustment (Score 16) & \textbf{20\%} & \textbf{20\%} & \textbf{30\%} & 0\% \\ \midrule
        \textbf{Stacking} & 5.1 & Shift garment to the left (Score 17) & 0\% & 0\% & 0\% & 0\% \\
         & 5.2 & Relocate garment to the right (Score 18) & \textbf{5\%} & \textbf{5\%} & 0\% & 0\% \\
         & 5.3 & Stack current garment onto pile (Score 19) & 0\% & 0\% & 0\% & 0\% \\
         & 5.4 & Relocate the stack to the right (Score 20) & 0\% & 0\% & 0\% & 0\% \\ \midrule
        \multicolumn{3}{r}{Total Failure Frequency} & 100\% & 65\% & 50\% & 25\% \\
        \multicolumn{3}{r}{\textbf{Success Rate}} & \textbf{0\%} & \textbf{35\%} & \textbf{50\%} & \textbf{75\%} \\ \bottomrule
    \end{tabular}

    \vspace{1em} 

    \includegraphics[width=1.0\linewidth]{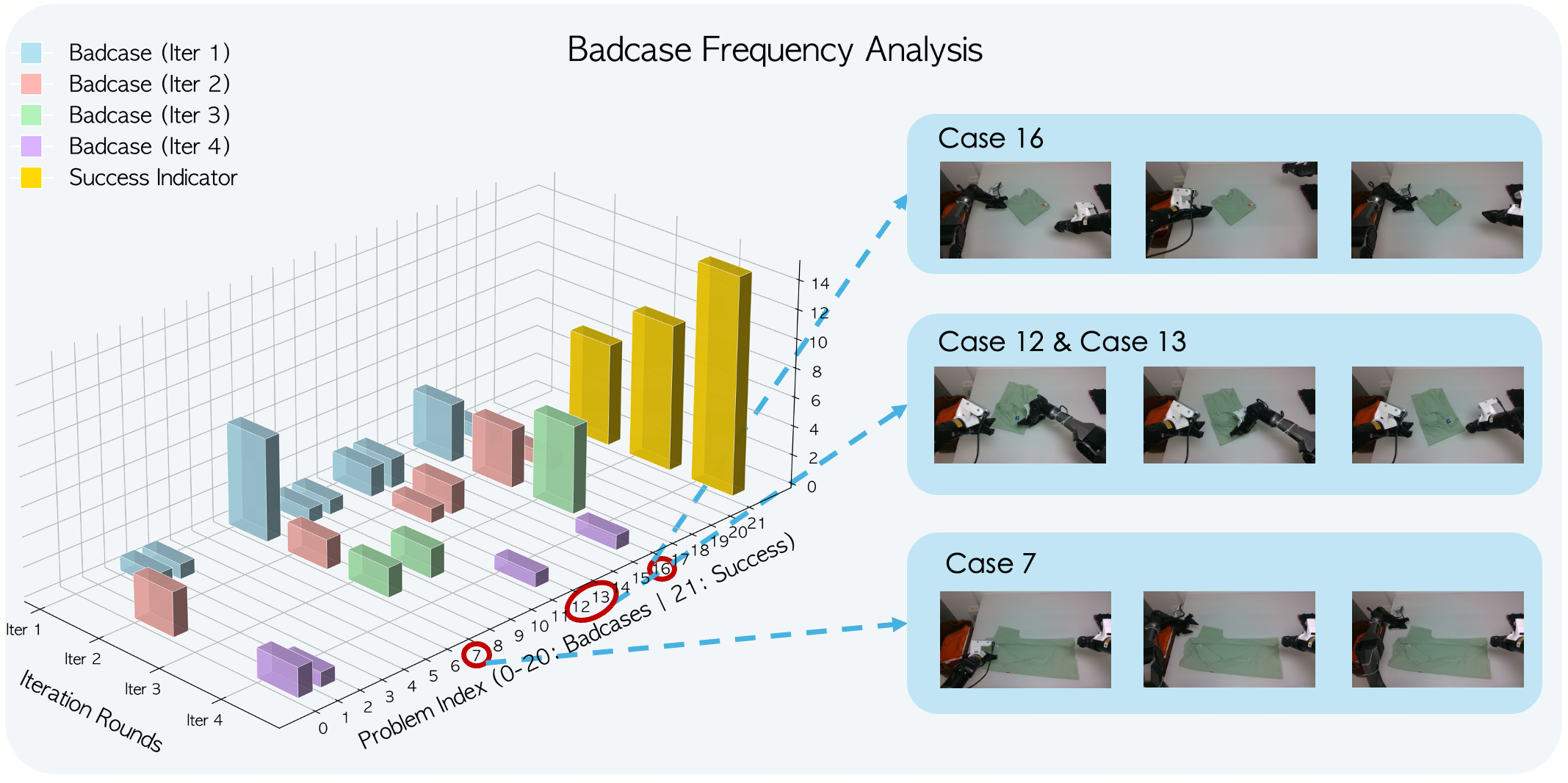} 
    \captionof{figure}{Badcase Frequency Analysis and Iterative Policy Improvement. The 3D histogram tracks the distribution of failure modes across four iteration rounds. By identifying high-frequency badcases (e.g., Cases 7, 12, 13, and 16), we implement targeted failure mode construction and state augmentation. The significant growth of the ``Success Indicator'' (yellow bars) demonstrates the effectiveness of targeted data collection in recovering from specific test-driven failures.}
    \label{fig:test_driven_data}

\end{table}

%

While proactive expansion broadens the state space, real-world deployment inevitably reveals specific, on-policy weaknesses. We address these through \textit{Test-Driven Failure Recovery}, a closed-loop optimization cycle.
%
As detailed in Table~\ref{tab:folding_corrected}, we categorize failure modes and their frequencies based on empirical deployment analysis.
Following the workflow illustrated in Fig.~\ref{fig:test_driven_data}, we then execute state expansion and targeted data collection specifically for high-frequency failures. 

 \begin{figure}[tb]
     \centering
     \includegraphics[width=1\linewidth]{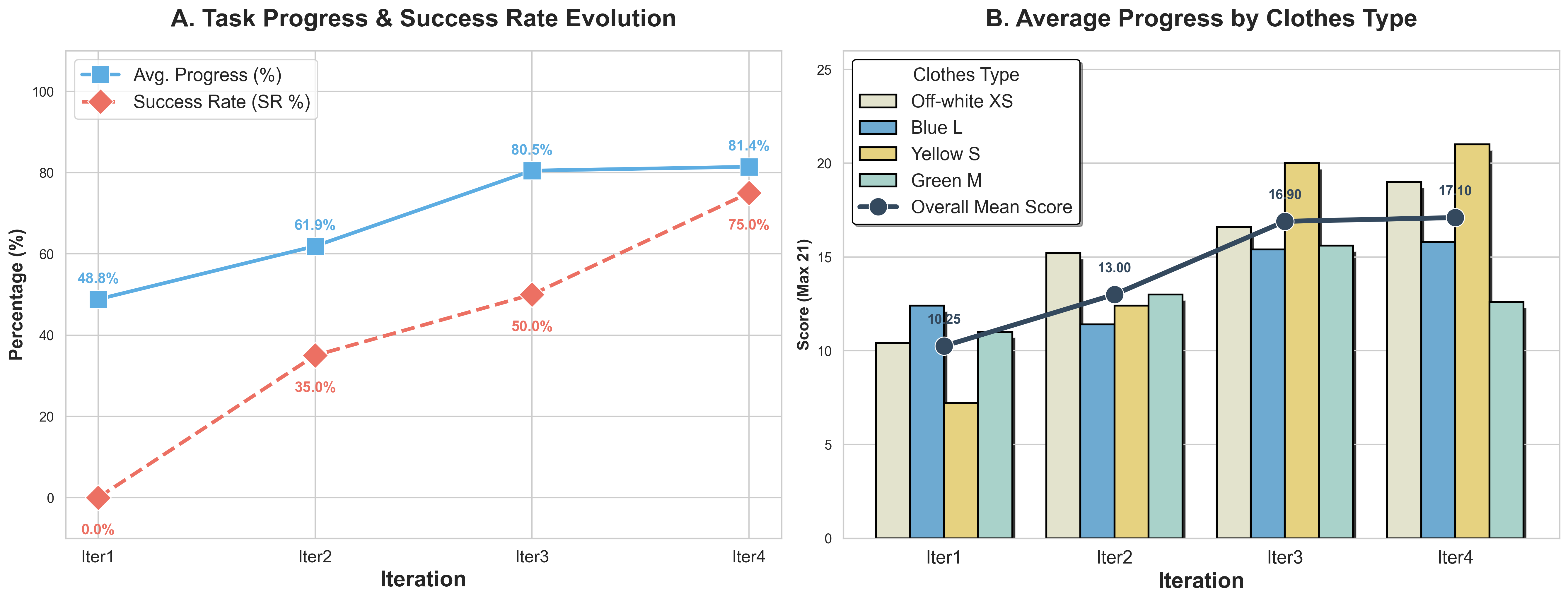}
     \caption{Quantitative Analysis of Iterative Training Results. Panel A highlights the positive correlation between iteration rounds and task completion metrics, specifically noting the significant jump in Success Rate during the final stages. Panel B illustrates the mean score progression for specific categories (e.g., Off-white XS, Yellow S), confirming a balanced performance gain across all test cases.}
     \label{fig:supp_realword_pic}
 \end{figure}

As shown in Fig.~\ref{fig:supp_realword_pic}, this strategy yields a consistent upward trend in performance. The Success Rate (SR) escalates from an initial 0\% to 75\% by Iteration 4.
A critical observation occurs in Iteration 4: while the \textit{average progress} increases marginally (80.5\% $\to$ 81.4\%), the \textit{Success Rate} surges significantly (50\% $\to$ 75\%). This disparity indicates that the final iteration successfully resolved "last-mile" long-tail failure modes (e.g., Case 7 and Case 16 in Fig.~\ref{fig:test_driven_data} ). Although these failures have a minor impact on the progress score (as the robot completes most of the task), correcting them is decisive for task completion.


Furthermore, we observe substantial gains in generalization (Panel B). 
Specifically, the "Yellow S" garment category, which performed poorly in Iteration 1, evolved into a top-performing class by Iteration 4. This confirms that our targeted data collection effectively bridges generalization gaps, compensating for initial performance disparities across diverse object variations.


\subsection{Data Quality Assurance and Curation}
Beyond data scale, we identify that the intrinsic quality of demonstrations is paramount for stable policy learning. We implement a rigorous data quality assurance protocol centered on the following four pillars:

\begin{itemize}[leftmargin=*]
    \item \textbf{Data Self-containment and Standardization:} To ensure geometric integrity and prevent systemic ambiguities (e.g., quaternion conventions), all trajectories are encapsulated using \textit{Protobuf} definitions. Each data package is self-describing, incorporating essential hardware metadata including URDF models and firmware versions. We strictly enforce that the end-effector (EE) pose link is explicitly defined within the URDF, with every kinematic link maintaining a rigid parent-child frame relationship to ensure transform consistency.
    
    \item \textbf{Visual Consistency Verification:} All collected demonstrations undergo a mandatory validation phase via real-robot replay. By re-projecting the EE trajectory onto the image plane using calibrated intrinsic and extrinsic camera parameters (as shown in Fig.~\ref{fig:data_quality_inspection}), we perform a pixel-level consistency check. This process allows for the systematic exclusion of invalid samples caused by calibration drifts, temporal desynchronization, or teleoperation errors.
    
    \item \textbf{Temporal Fidelity and Causal Filtering:} To mitigate \textit{causal confusion} in state-action mapping, we employ automated scripts to prune non-informative static frames. However, we introduce a context-aware filtering heuristic: static joint states are preserved if they coincide with critical task phases—such as waiting for a container to fill or a conveyor to deliver an object—as identified through synchronized vision and task-level metadata. This ensures that purposeful pauses are distinguished from redundant idle data.
    
    \item \textbf{Multi-embodiment Consistency:} To minimize the domain gap introduced by mechanical variances, validated episodes are replayed across multiple physical robot entities. This cross-robot validation ensures that the learned operational logic remains invariant to specific hardware instances, effectively isolating and rectifying discrepancies stemming from subtle calibration errors or mechanical tolerances.
\end{itemize}

\subsection{Failure Mode Analysis}
Despite the iterative refinements leading to version 4.0, certain "long-tail" failure modes persist in the complex cloth-folding task. We categorize and analyze these persistent challenges as follows:

\begin{itemize}[leftmargin=*]
    \item \textbf{Progress Reversion:} A primary challenge in manipulating thin-shell deformable objects is the high-dimensional state space and the existence of irreversible failure modes. We observed cases where the policy, unable to recover from a tangled configuration through local adjustments, caused the task progress to revert entirely to the initial flattening phase. This "reset-to-zero" behavior highlights the lack of robust mid-level recovery strategies for highly disordered states.
    
    \item \textbf{State Confusion:} During the folding sequence, the visual appearance of partially folded garments often resembles that of a crumpled state. This perceptual ambiguity occasionally causes the policy to misidentify the current task stage, leading to "state oscillation" where the robot re-executes a flattening action in the middle of a folding trajectory.
    
    \item \textbf{Insufficient Clothes Generalization:} The physical properties of garments significantly influence manipulation success. For instance, low-friction materials like silk frequently lead to "empty grasps" or slippage. The model struggles to generalize across more diverse sizes and intricate designs, although performance remains consistent on standard solid-colored T-shirts.
    
    \item \textbf{Low Flattening Efficiency:} While our Standard Operating Procedure (SOP) utilizes repetitive primitives for flattening, human demonstrators often employ "implicit heuristics"—such as specific tensioning forces and strategic grasp point selection—that are difficult to capture through limited data. Consequently, the model exhibits lower efficiency in resolving highly disordered states compared to the nuanced techniques demonstrated by human experts.
\end{itemize}

\clearpage

\section{Details Result for Simulation Benchmarks}
\label{supp_sec:exp_details}

In this section, we present detailed evaluation metrics for our model across three simulation benchmarks: RoboTwin-2.0 (Table~\ref{tab:robotwin_detail_qw} and Table~\ref{tab:robotwin_detail_gd}), 
and LIBERO-Plus (Table~\ref{tab:supp_liberoplus_qw} and Table~\ref{tab:supp_liberoplus_gd}).

\begin{table}[h]
\centering
\caption{Detailed results of \modelname -GD on the RoboTwin-2.0 benchmark.}
\label{tab:robotwin_detail_gd}
\small
\setlength{\tabcolsep}{4pt}

\begin{tabular}{lcc|lcc|lcc}
\toprule
\multicolumn{1}{c}{Task} & Clean & Rand. & \multicolumn{1}{c}{Task} & Clean & Rand. & \multicolumn{1}{c}{Task} & Clean & Rand. \\
\midrule
Adjust Bottle & 100.0 & 100.0 & Open Microwave & 99.0 & 98.0 & Place Object Stand & 92.0 & 92.9 \\
Beat Block Hammer & 98.0 & 91.0 & Pick Diverse Bottles & 86.0 & 87.0 & Place Phone Stand & 92.0 & 98.0 \\
Blocks Ranking RGB & 100.0 & 98.0 & Pick Dual Bottles & 98.0 & 97.0 & Place Shoe & 99.0 & 98.0 \\
Blocks Ranking Size & 85.0 & 87.0 & Place A2B Left & 70.0 & 76.8 & Press Stapler & 96.0 & 90.0 \\
Click Alarmclock & 99.0 & 100.0 & Place A2B Right & 72.0 & 66.0 & Put Bottles Dustbin & 93.0 & 97.0 \\
Click Bell & 100.0 & 100.0 & Place Bread Basket & 97.0 & 95.0 & Put Object Cabinet & 92.0 & 87.0 \\
Dump Bin Bigbin & 92.0 & 98.0 & Place Bread Skillet & 82.0 & 86.0 & Rotate QRcode & 90.0 & 89.0 \\
Grab Roller & 100.0 & 100.0 & Place Burger Fries & 99.0 & 99.0 & Scan Object & 85.0 & 80.0 \\
Handover Block & 96.0 & 93.0 & Place Can Basket & 89.0 & 80.0 & Shake Horizontally & 100.0 & 100.0 \\
Handover Mic & 100.0 & 100.0 & Place Cans Plasticbox & 100.0 & 99.0 & Shake Bottle & 100.0 & 100.0 \\
Hanging Mug & 48.0 & 45.0 & Place Container Plate & 99.0 & 100.0 & Stack Blocks Three & 96.0 & 98.0 \\
Lift Pot & 100.0 & 100.0 & Place Dual Shoes & 92.0 & 95.0 & Stack Blocks Two & 100.0 & 97.0 \\
Move Can Pot & 100.0 & 99.0 & Place Empty Cup & 100.0 & 100.0 & Stack Bowls Three & 92.0 & 81.0 \\
Move Pillbottle Pad & 96.0 & 97.0 & Place Fan & 90.0 & 92.0 & Stack Bowls Two & 95.0 & 96.0 \\
Move Playingcard Away & 92.0 & 93.0 & Place Mouse Pad & 76.0 & 76.0 & Stamp Seal & 75.0 & 76.0 \\
Move Stapler Pad & 74.0 & 77.0 & Place Object Basket & 87.0 & 81.0 & Turn Switch & 70.0 & 67.0 \\
\cline{7-9} 
Open Laptop & 95.0 & 100.0 & Place Object Scale & 87.0 & 85.0 & \textbf{Average} & \textbf{91.3} & \textbf{90.8} \\
\bottomrule
\end{tabular}
\end{table}

\begin{table}[h]
\centering
\caption{Detailed results of \modelname -QW on the RoboTwin-2.0 benchmark.}
\label{tab:robotwin_detail_qw}
\small
\setlength{\tabcolsep}{4pt}

\begin{tabular}{lcc|lcc|lcc}
\toprule
\multicolumn{1}{c}{Task} & Clean & Rand. & \multicolumn{1}{c}{Task} & Clean & Rand. & \multicolumn{1}{c}{Task} & Clean & Rand. \\
\midrule
Adjust Bottle & 100.0 & 99.0 & Open Microwave & 97.0 & 99.0 & Place Object Stand & 94.0 & 94.0 \\
Beat Block Hammer & 92.0 & 91.0 & Pick Diverse Bottles & 89.0 & 87.0 & Place Phone Stand & 94.0 & 98.0 \\
Blocks Ranking RGB & 98.0 & 96.0 & Pick Dual Bottles & 98.0 & 96.0 & Place Shoe & 100.0 & 100.0 \\
Blocks Ranking Size & 81.0 & 91.0 & Place A2B Left & 77.0 & 81.0 & Press Stapler & 91.0 & 90.0 \\
Click Alarmclock & 99.0 & 99.0 & Place A2B Right & 74.0 & 73.0 & Put Bottles Dustbin & 97.0 & 98.0 \\
Click Bell & 97.0 & 100.0 & Place Bread Basket & 93.0 & 90.0 & Put Object Cabinet & 89.0 & 85.0 \\
Dump Bin Bigbin & 88.0 & 92.0 & Place Bread Skillet & 91.0 & 89.0 & Rotate QRcode & 89.0 & 95.0 \\
Grab Roller & 100.0 & 100.0 & Place Burger Fries & 98.0 & 95.0 & Scan Object & 84.0 & 84.0 \\
Handover Block & 100.0 & 90.0 & Place Can Basket & 79.0 & 90.0 & Shake Horizontally & 100.0 & 98.0 \\
Handover Mic & 100.0 & 100.0 & Place Cans Plasticbox & 100.0 & 100.0 & Shake Bottle & 100.0 & 99.0 \\
Hanging Mug & 55.0 & 52.0 & Place Container Plate & 100.0 & 99.0 & Stack Blocks Three & 93.0 & 94.0 \\
Lift Pot & 100.0 & 100.0 & Place Dual Shoes & 96.0 & 96.0 & Stack Blocks Two & 100.0 & 99.0 \\
Move Can Pot & 99.0 & 97.0 & Place Empty Cup & 100.0 & 100.0 & Stack Bowls Three & 88.0 & 88.0 \\
Move Pillbottle Pad & 95.0 & 91.0 & Place Fan & 96.0 & 92.0 & Stack Bowls Two & 97.0 & 96.0 \\
Move Playingcard Away & 88.0 & 96.0 & Place Mouse Pad & 82.0 & 83.0 & Stamp Seal & 77.0 & 85.0 \\
Move Stapler Pad & 73.0 & 84.0 & Place Object Basket & 92.0 & 90.0 & Turn Switch & 89.0 & 79.0 \\
\cline{7-9} 
Open Laptop & 97.0 & 98.0 & Place Object Scale & 91.0 & 97.0 & \textbf{Average} & \textbf{91.9} & \textbf{92.3} \\
\bottomrule
\end{tabular}
\end{table}

\begin{table}[t]
\centering
\caption{Detailed results of \modelname -QW on the LIBERO-PLUS benchmark}
\label{tab:supp_liberoplus_qw}
\small

\begin{tabular}{l|c|cccccccc}
\toprule
 & Original & Camera & Robot & Language & Light & Background & Noise & Layout & Total \\
\midrule
Spatial & 97.2 & 84.0 & 60.6 & 77.7 & 96.6 & 96.5 & 87.2 & 88.1 & 83.6 \\
Object  & 99.6 & 87.4 & 47.0 & 74.2 & 99.7 & 96.8 & 89.3 & 81.1 & 80.9 \\
Goal    & 97.6 & 48.0 & 36.9 & 49.3 & 91.4 & 89.0 & 54.4 & 64.5 & 59.2 \\
Long    & 95.2 & 48.2  & 53.2 & 64.0 & 91.6 & 91.7 & 62.6 & 81.1 & 67.7 \\
\midrule
Avg     & 97.4 & 66.3 & 49.0 & 65.9 & 94.9 & 93.3 & 73.1 & 78.2 & 72.6 \\
\bottomrule
\end{tabular}
\end{table}

\begin{table}[t]
\centering
\caption{Detailed results of \modelname -GD on the LIBERO-PLUS benchmark}
\label{tab:supp_liberoplus_gd}
\small

\begin{tabular}{l|c|cccccccc}
\toprule
 & Original & Camera & Robot & Language & Light & Background & Noise & Layout & Total \\
\midrule
Spatial & 97.8 & 73.9 & 81.4 & 96.7 & 94.2 & 91.4 & 67.2 & 92.5 & 85.1 \\
Object  & 98.2 & 87.1 & 51.5 & 95.5 & 98.7 & 96.4 & 86.3 & 80.9 & 83.8 \\
Goal    & 95.2 & 45.1 & 47.4 & 42.9 & 75.3 & 86.8 & 59.4 & 64.7 & 58.2 \\
Long    & 95.6 & 57.5  & 55.5 & 83.0 & 83.2 & 87.5 & 54.8 & 82.0 & 69.9 \\
\midrule
Avg     & 96.7 & 65.5 & 58.2 & 78.7 & 88.1 & 90.3 & 66.9 & 79.5 & 74.0 \\
\bottomrule
\end{tabular}
\end{table}

\end{document}